\documentclass[twoside]{article}

\usepackage[accepted]{aistats2020}
\usepackage[round]{natbib}

\usepackage[utf8]{inputenc} 
\usepackage[T1]{fontenc}    
\usepackage[draft]{hyperref}       
\usepackage{url}            
\usepackage{booktabs}       
\usepackage{amsfonts}       
\usepackage{nicefrac}       
\usepackage{microtype}      
\usepackage{graphicx}
\usepackage{amsmath}
\usepackage{amsthm}
\usepackage{xcolor}
\usepackage[english]{babel}
\usepackage{epstopdf}

\makeatletter
\newtheorem*{rep@theorem}{\rep@title}
\newcommand{\newreptheorem}[2]{%
\newenvironment{rep#1}[1]{%
 \def\rep@title{#2 \ref{##1}}%
 \begin{rep@theorem}}%
 {\end{rep@theorem}}}
\makeatother

\newtheorem{lemma}{Lemma}
\newreptheorem{lemma}{Lemma}

\begin{document}

\twocolumn[

\aistatstitle{Deontological Ethics By Monotonicity Shape Constraints}

\aistatsauthor{ Serena Wang \And Maya Gupta }
\aistatsaddress{Google Research \And Google Research}
]

\begin{abstract}
We demonstrate how easy it is for modern machine-learned systems to violate common deontological ethical principles and social norms such as ``favor the less fortunate,'' and ``do not penalize good attributes.'' We propose that in some cases such ethical principles can be incorporated into a machine-learned model by adding shape constraints that constrain the model to respond only positively to relevant inputs. We analyze the relationship between these deontological constraints that act on individuals and the consequentialist group-based fairness goals of one-sided statistical parity and equal opportunity.  This strategy works with sensitive attributes that are Boolean or real-valued such as income and age, and can help produce more responsible and trustworthy AI. 
\end{abstract}

\section{Introduction}
As the use of machine-learned (ML) models broadens, strategies are sought to ensure machine-learned systems behave \emph{responsibly}, and are  \emph{ethical} and \emph{fair}. There may be many different reasonable and conflicting ethical stances for a given problem \citep{Haidt:2013,EthicsInTheRealWorld, Atlas:2018,Corbett-Davies:2018,binns:2018}. Thus, no single strategy for making machine learning fairer is likely to be sufficient. 

\begin{figure}[hb!]
\centering
\begin{tabular}{c}
\includegraphics[width=0.45\textwidth]{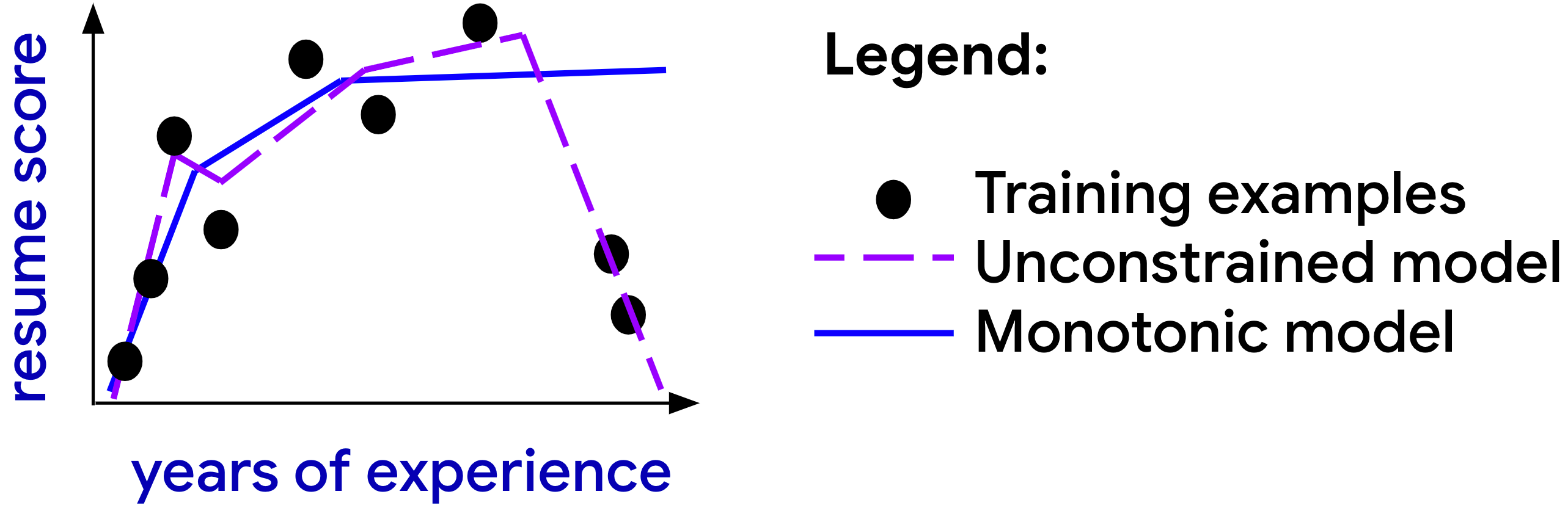} \\
\end{tabular}
\caption{Toy example showing how a monotonicity constraint can protect against unfair penalization.}
\label{fig:age} 
\end{figure}

In this paper, we show that nonlinear machine learning can easily produce trained models that violate social norms or ethics that can be described as:  \emph{certain inputs should not have a negative effect on an outcome}. For example, the toy example in Fig. \ref{fig:age} shows a nonlinear 1-d model trained to predict how highly a resume will be scored based on the candidate's years of experience. The best fit model (purple dashed line) sometimes penalizes candidates for having \emph{more} job experience. For candidates with many years of experience, the dashed model picked up the age discrimination in the biased training samples. In addition, the dashed model also penalizes job experience in the lower-end of years of experience simply due to overfitting. Such unfair responses are 
very much a danger in sparse regions of a feature space with modern over-parameterized nonlinear models.

Fig. \ref{fig:age} also illustrates our proposed solution (blue solid line): train the model with monotonicity shape constraints to guarantee that the model can only use job experience as positive evidence. This is a \textit{deontological} solution, and thus differs from many existing mathematical expressions of fairness that are \emph{consequentialist} and \emph{statistical} \citep{Friedler:2019, hardt:2016,binns:2018,Sandvig:2016}. Both types of goals may be of interest to a practitioner \citep{Sandvig:2016}, and we explore how they relate theoretically and experimentally. 

The main contributions of this paper are: \emph{(i)} identifying that violations of monotonicity in ML models may pose significant ethical issues and risk of societal harm, \emph{(ii)} demonstrating experimentally that these violations can easily occur in practice with nonlinear machine learned models, \emph{(iii)} showing that existing shape constrained ML can be used effectively to ameliorate such problems, and \emph{(iv)} theoretically and empirically analyzing how the proposed monotonicity constraints relate to statistical fairness notions. 




\begin{table}[ht!]
\centering
\begin{tabular}{ll|ll}
\hline
Model type & \# shape  & Train  & Test\\
& constraints & Acc. & Acc. \\
\hline
GAM & 0  & $94.93\%$ & $94.89\%$\\
GAM & $2$  & $94.90\%$ & $\mathbf{94.97}\%$ \\
DNN & 0  & $94.97\%$ & $94.89\%$ \\
GBT & 0  & $\mathbf{95.04}\%$ & $94.80\%$ \\
\hline
\end{tabular}
\caption{Law School Admissions Experiment Results. Two monotonicity constraints ensure that individuals aren't penalized for higher GPA or LSAT score.} 
\label{tab:law}
\end{table} 

\section{Illustrating Unfair Penalization}
\begin{figure*}[h!]
\centering
\begin{tabular}{llll}
(a) Neural Network  & (b) Grad. Boosted Trees  & (c) Monotonic GAM & (d) GAM \\
\includegraphics[width=0.229\textwidth]{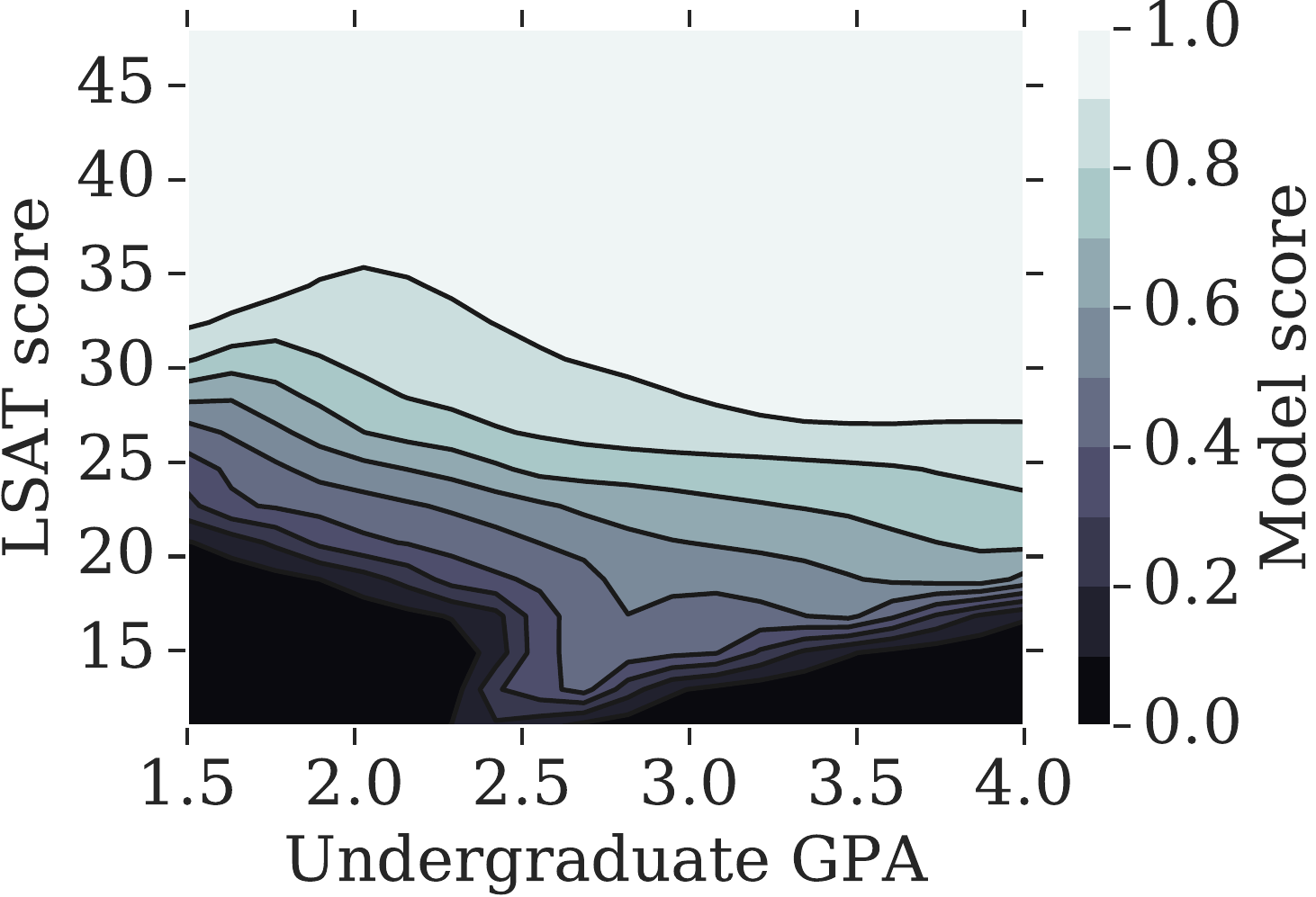} & 
\includegraphics[width=0.229\textwidth]{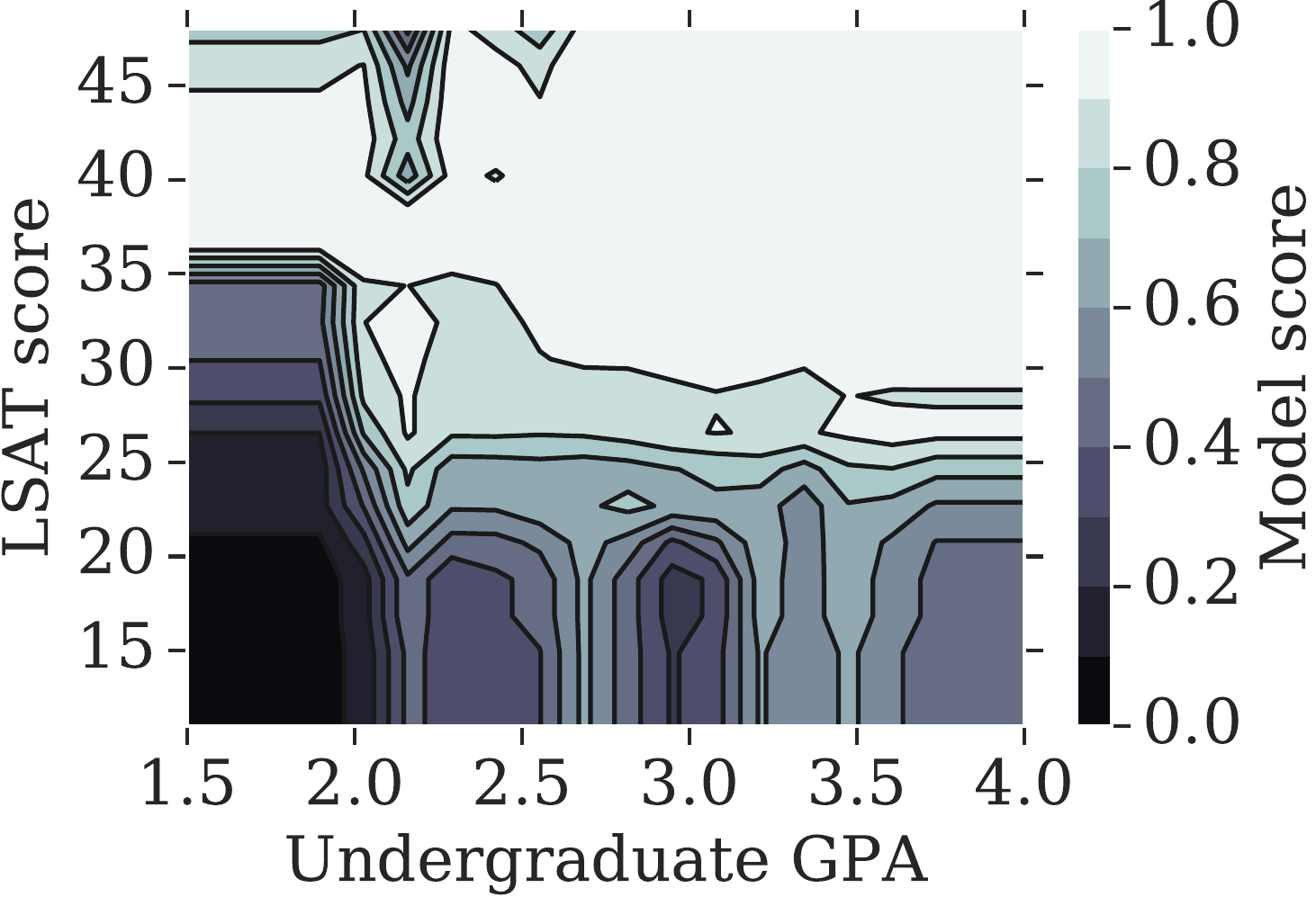} &
\includegraphics[width=0.229\textwidth]{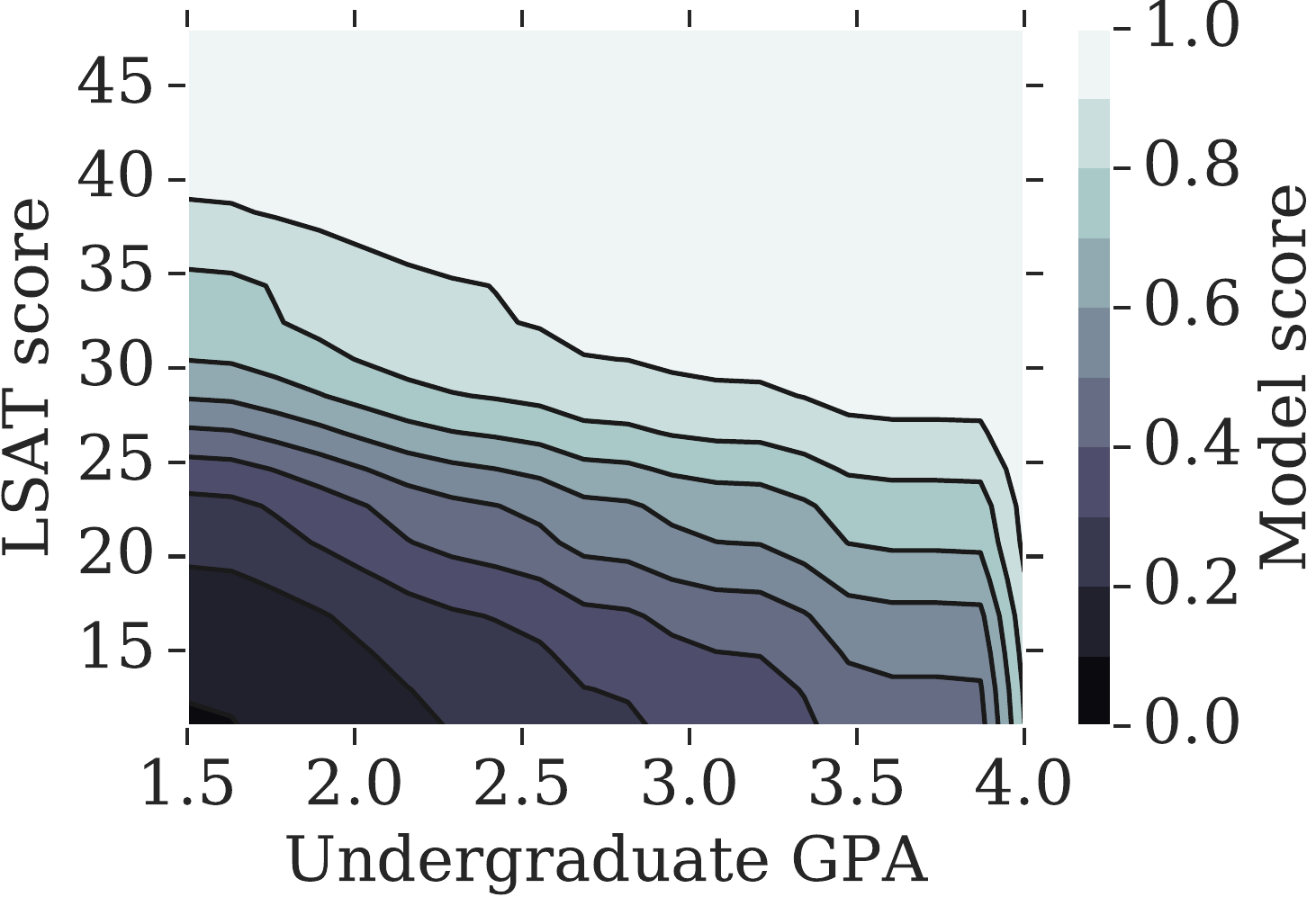} &
\includegraphics[width=0.229\textwidth]{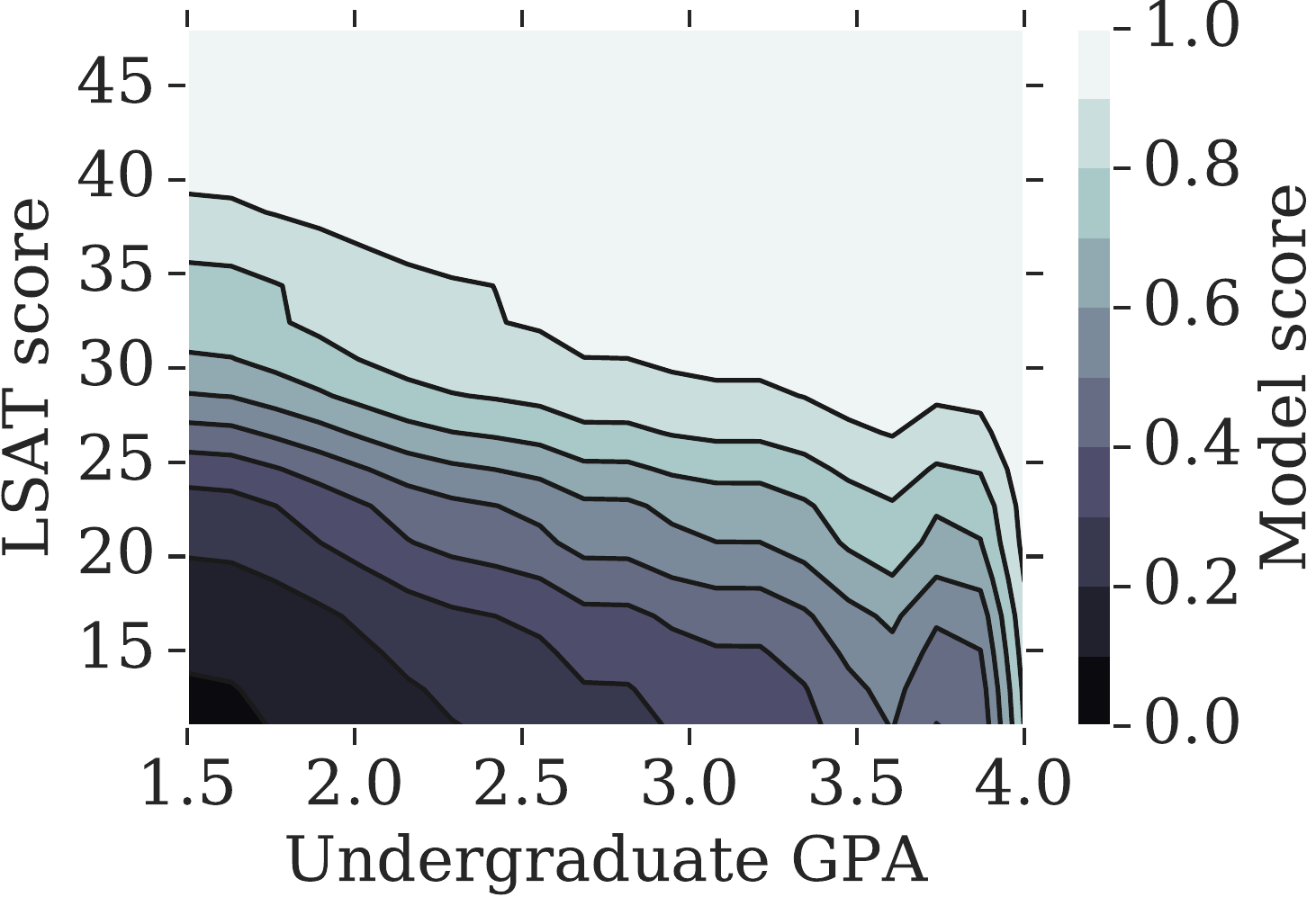} \\
\end{tabular}
\caption{Real data example predicting if a student will bass the bar.   Predictions by a neural network \textit{(a)} and gradient boosted trees \textit{(b)} penalize some students if they increase their GPA or LSAT score. We trained a generalized additive model (GAM) with constraints to be monotonically increasing in both GPA and LSAT score \textit{(c)}. The same GAM model trained \emph{without} monotonicity constraints also violates monotonicity in GPA \textit{(d)}.}
\label{fig:law_models} 
\end{figure*}

To further illustrate the potential for machine-learning to produce objectionable models, consider the \textit{Law School Admissions} dataset \citep{Wightman:1998}. Suppose this data is used to predict whether a person would pass the bar exam based on their LSAT score and undergraduate GPA, and that the classifier's score was used to guide law school admissions or scholarships. 
We trained a standard two-layer deep neural network (DNN) (more experimental details in Sec. \ref{sec:experiments}) and show the model's output for each possible input in 
Fig. \ref{fig:law_models}. The DNN sometimes penalizes people for having a higher GPA:  for example, with an LSAT score of 15, the DNN rewards students with a lower 2.7 GPA over students who earned a higher 3.5 GPA. Similarly, if a student has a GPA of 2.5, the DNN gives that student a higher score for scoring 10 on the LSAT than if they had scored 15. Thus this model violates merit-based social norms and the ``best-qualified'' ethical principle \citep{HunterSchmidt:1976} (we also acknowledge that the very use of standardized test scores or GPA for allocating goods may raise other ethical issues \citep{HunterSchmidt:1976}). Training with gradient boosted trees has the same problem: the model penalizes some people for raising their GPA or LSAT score.   

Each of the two-dimensional models shown in Fig. \ref{fig:law_models} were trained on 19,064 training examples, which may sound like plenty of data to learn a model on two inputs, but the non-uniform distribution of the training data means that some regions of the feature space were sparse and the model may have overfit (for a density plot, see Fig. \ref{fig:law_data} in the Appendix).  These monotonicity violations can also occur from legitimate, clean data. To see this, note that how hard it is to get a good GPA can vary greatly between schools, and imagine as an extreme example that there was a large university that simply gave every student a 2.7 GPA, which could cause some of the problems seen in this model. 


Our proposal is to train the model with monotonicity constraints. An example monotonic model is shown in \textit{(c)} of Fig. \ref{fig:law_models}. It is a generalized additive model (GAM) trained with the  constraints that it never penalizes higher GPA's for any LSAT score, and that it never penalizes higher LSAT scores for any GPA. Training the same GAM \emph{without} monotonicity constraints is shown in the lower right, and also produces an objectionable model.  The test accuracies of these four models are given in Table \ref{tab:law} are similar (more experimental details in Sec. \ref{sec:experiments}).

\section{Monotonicity Fairness Constraints}\label{sec:monotonic}
The motivating examples in Figures 1 and 2 focused on the concern of \emph{unfair penalization} -- that there may be inputs that a responsible model may reward but should never penalize. A second ethical pattern we consider is to \emph{favor the less fortunate} -- there may be inputs that help us identify the less fortunate and favor them if there are no other relevant differences.  Policies favoring the less fortunate have been well studied in economics \citep{Coate:1993}. As shown in Figures 1 and 2, we propose addressing these two principles by constraining the ML model to only respond positively to relevant inputs if all other inputs are fixed.   

Throughout, we focus on ML models that produce a score that is used to determine some benefit, such as a better credit rating or a scholarship. Specifically, consider a learned model $f(x, z)$ where  $x \in \mathbb{R}^D$ is a $D$-dimensional feature vector, and $z \in \mathbb{R}$ is another input that is of ethical interest (such as age or income).  The model $f(x,z)$  satisfies a \emph{positive monotonicity shape constraint} \citep{Groeneboom:2014} with respect to $z$ if for any $\delta > 0$ and any choice of $x$ and $z$, $f(x,z + \delta) \geq f(x,z)$. If $f$ is differentiable with respect to $z$, this constraint is equivalent to non-negative slope $\frac{\partial(f(x,z))}{\partial z} \geq 0$ for any $x,  z$.  \emph{Strict monotonicity} replaces the $\geq$ sign above with $>$. Reverse definitions produce negative monotonicity constraints.  To apply a monotonicity shape constraint to a categorical feature, one can express each category as a Boolean feature indicating membership in the category.

As defined above, and as is standard in the shape constraints literature \citep{Groeneboom:2014}, our proposed monotonicity shape restrictions are \emph{ceterus paribus}: that is, $f(x)$ should behave monotonically with respect to increases in each protected feature $z$, but only \emph{when all other features} $x$ \emph{are held fixed}. Here we take the input features $x$ as given, but of course it is also important to have the best possible set of features $x$. 

\section{More Example Scenarios}
We further illustrate the critical importance and breadth of situations in which we need ML models to behave consistently with a society's norms or ethical policies.  Then in Section \ref{sec:experiments}, we show experimentally on two more problems that nonlinear ML models do not naturally pick up such norms and policies, and that training with monotonicity constraints can be used to incorporate desired policies.

\textbf{Crimes and Misdemeanors:} An ML system trained to determine fines for misdemeanors would likely be considered fairer if the fines were monotonically increasing with respect to the magnitude of the illegality, e.g. a larger fine for illegally parking one's car for longer, or for exceeding the speed limit by more \citep{Corrections}. Also, in many societies it is expected that a juvenile will not be penalized more heavily than an adult for the same crime, all else equal \citep{Corrections}. Many societies prefer not to give harsher penalties to first-time offenders than to repeat offenders \citep{Corrections}. 

\textbf{Pay:} People generally feel it is unfair to get paid less for doing more of the same work \citep{Fehr}.  Suppose a model is trained to advise parents on how much to pay their babysitter. It may be desirable to workers if the recommended pay were constrained to be a monotonically increasing function of the number of hours worked, all else equal. Similarly, consider an app that uses an ML model to advise people on how much to tip their waiter in America. Such a model would be better aligned with American societal norms if it recommended larger tips for more expensive meals, and if it recommended higher tip percentages for more upscale establishments, all else equal \citep{Azar:2004}. 

\textbf{Medical Triage:} In some medical contexts, it is considered more ethical or more responsible to prioritize patients based on their risk or neediness \citep{Triage:2007}. For example, the United States Transplant Board has a policy of giving a sicker person a higher score to receive a transplant for ethical reasons \citep{WashingtonPost}. Similarly an emergency room scoring patients for prioritization may require that patients that have waited longer are treated first, if all other relevant characteristics are equal \citep{Triage:2007}.  More generally, \emph{first-come first-served} is a common principle underpinning civil society \citep{FCFS,SpontaneousOrder}.  


\section{Related Work in Ethical ML}\label{sec:relatedWork}
This work fits into the broader literature on \emph{machine ethics} \citep{AndersonSurvey:2007, Moor:2006}. Ethics itself is a broad field of philosophy that encompasses many questions, and there may be many different reasonable ethical stances and criteria for a given problem \citep{binns:2018,Thiroux:Ethics,Haidt:2013,Atlas:2018,Corbett-Davies:2018,EthicsInTheRealWorld,Gruetzemacher:2018,Corrections,Rawls:71}. 



\textbf{Deontological:} Most recent work in machine learning fairness has been \emph{consequentialist}: focusing on ensuring that machine-learned models deliver statistically-similar performance for different groups \citep{binns:2018,Zafar:2015,Goh:2016,hardt:2016, Zafar:2017,Donini:2018,Agarwal:2018,Goel:2018,Cotter:2019,Cotter:2019b}. Those efforts solve a different important problem than the one we target here, and are complementary to this proposal. This proposal is instead \emph{deontological}, in that it enables people to impose rules on how the model can respond to inputs, producing an \emph{implicit ethical agent} in the terminology of Moor \citep{Moor:2006}. Sandvig et al.  recently called for more research into deontological algorithms, noting that, ``Applied ethics in real-world settings typically incorporates both rule-based and consequences-based reasoning'' \citep{Sandvig:2016}. 


\textbf{Monotonicity and Fairness:}
Concurrently, \citet{Cole:2019} also recognized the importance of monotonicity in a fairness context. That work differs in framing: they use monotonicity to reduce unfair \textit{resentment}. Our work further differs in its theoretical results and comparisons.

\textbf{Asymmetric:} Another fairness principle is that \emph{similar individuals should receive similar treatment} \citep{dwork:2012}; while also important, that principle aims for equal outcomes for two examples, whereas this proposal aims to make sure that any unequal treatment is unequal in the appropriate direction, such as \emph{favor the less fortunate}.  

\textbf{Counterfactual Fairness:}  Counterfactual fairness \citep{Kusner:2017,Pearl:2016} says that changing a protected attribute $A$ while holding things not causally dependent on $A$ constant will not change the distribution of the model output. This is similar to the definition of monotonicity in section \ref{sec:monotonic}, but is focused on treating certain cases the \emph{same} rather than prefering one case to another.


\textbf{Continuous Sensitive Features:}  The proposed monotonicity constraints handles real-valued attributes natively. Other recent efforts have also been devised to handle continuous protected attributes, like age, for consequentialist fairness goals \citep{Raff:2018,Kearns:2018,Komiyama:2018}. 


\textbf{Fair Ranking and Fair Regression:} 
Monotonicity constraints can be applied to ranking and regression models.  Other notions of ethics have also been considered for ranking models \citep{Berk:2017,Zehlike:2017,Celis:2018,SIRpairwise:2019,SinghJ18} and regression \citep{Komiyama:2018,perez2017fair,Berk:2017,Agarwal:2019}.

\section{Related Work in Training Models with Monotonicity Constraints}\label{app:engineer}
Monotonicity shape constraints have long been used to capture prior knowledge and regularize estimation problems to improve a model's generalization to new test examples \citep{Barlow:1972,Chetverikov:2018,BenDavid:92,Archer:93,Sill:97,kotlowski2009rule,Groeneboom:2014,GuptaEtAl:2016,canini:2016,You:2017,Lafferty:2018}. In this paper, we point out that monotonicity shape constraints can and should \emph{also} be used to ensure that machine-learned models behave consistently with societal norms and prima facie duties \citep{Ross:Duties} that can be expressed as monotonic relationships.

Unlike its use for regularization, applying monotonicity constraints to impose ethical principles may actually hurt test accuracy if the training and test data is biased (as in the toy example of Fig. 1). However, in all three of our real-data experiments the test accuracy was little changed by adding these constraints (see Section \ref{sec:experiments} and Fig. 2 and Appendix \ref{app:law}). 


Constraining multi-dimensional models to obey monotonic shape constraints has been shown to work for a variety of function classes, including neural networks \citep{Archer:93,Sill:97} and trees \citep{BenDavid:92,kotlowski2009rule,Lafferty:2018}. 

 
Our experiments use the open-source TensorFlow Lattice 2.0 package \citep{TFLatticeBlogPost},
which enables training GAMs and lattice models with monotonicity constraints \citep{GuptaEtAl:2016,canini:2016,You:2017}. 


\section{Relationship to Statistical Fairness} \label{sec:theory} 



In this section we analyze how the proposed deontological monotonicity constraints interact with consequentialist statistical fairness goals that are based on aggregate outcomes. For example, suppose a national funding agency enforces that poorer schools are funded more often as richer schools on average (we will call this \emph{one-sided statistical parity}). By only considering the national average funding rates, richer schools may still get funded more often than poorer schools in some states. In contrast, if the state were the only input $X$, then a deonotological monotonicity constraint would guarantee that for each state, poorer schools would get a higher funding rate than richer schools.



\citet{hardt:2016} showed that any \textit{oblivious} statistical fairness measure that doesn't depend on $X$ or the function form $f(X,Z)$ can fail to identify forms of discrimination. Popular statistical notions like \textit{statistical parity} \citep{dwork:2012} and \textit{equal opportunity} \citep{hardt:2016} are oblivious, whereas monotonicity is not.
In the next section we show specifically that satisfying \textit{one-sided statistical parity} does not imply monotonicity, but that satisfying monotonicity can impact and bound one-sided statistical parity and equal opportunity (proofs in Appendix).





\subsection{Bounds on One-sided Statistical Parity}
\emph{Statistical parity} is a well known measure of fairness for ML models  \citep{dwork:2012,Zafar:2015,Cotter:2019b}. Consider a model $f(X, Z)$ that takes as input a random feature vector $X \in \mathbb{R}^D$ and a random protected attribute $Z$, which may be categorical or real-valued. Suppose the  model $f$ outputs a real-valued score (we treat the special case of classifiers next). Then \emph{statistical parity} requires $E[f(X, Z) | Z = j] = E[f(X, Z) | Z = k]$ for any $j,k$. Because we focus on asymmetric goals like favoring the less fortunate, we consider \textit{one-sided} statistical parity: $E[f(X, Z) | Z = j] \leq E[f(X, Z) | Z = k]$ for any $j \leq k$.

We show in Appendix \ref{app:moncounter}  that due to Simpson's paradox \citep{Bickel:75}, that a monotonicity constraint on $Z$ is not sufficient to guarantee one-sided statistical parity with respect to $Z$. However, monotonicity does imply a \textit{bound} on the one-sided statistical parity violation: in Lemma \ref{lem:measure}, we show that if $f$ is monotonic with respect to $Z$, then the one-sided statistical parity violation between $Z =j$ and $Z=k$ will be bounded by the maximum density ratio between the two groups.

\begin{figure}[h!]
\centering
\begin{tabular}{c}
\includegraphics[width=0.35\textwidth]{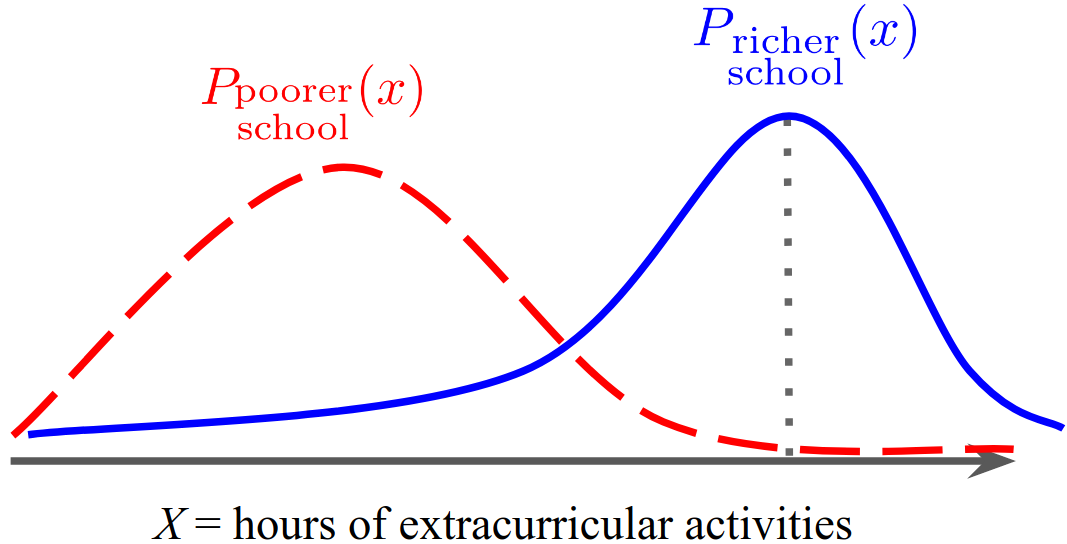} \\
\end{tabular}
\caption{Illustration of Lemma 1. The dotted line shows the $X$ value achieving the max density ratio $C$.}
\label{fig:theory} 
\end{figure}

\begin{lemma}\label{lem:measure} 
Let $(\Omega, \mathcal{F})$ be a measurable space with a regular conditional probability property, and let $X: \Omega \to \mathbb{R}^D$, $Z: \Omega \to \mathbb{R}$ be $\mathcal{F}$-measurable random variables. Suppose $P_j$ and $P_k$ are $\sigma$-finite probability measures on $(\Omega, \mathcal{F})$, where $P_j$ denotes the conditional probability measure of $X$ given that $Z = j$, and $P_k$ denote the same for $Z = k$, and $P_j$ is absolutely continuous with respect to $P_k$. Let $f: \mathbb{R}^D \times \mathbb{R} \to \mathbb{R}$ be defined as in Section \ref{sec:monotonic}, and $f(x,z) \geq 0$ for all $x \in \mathbb{R}^D$, $z \in \mathbb{R}$. If the function $f$ satisfies monotonicity in the second argument such that for $j \leq k$, $f(x,j) \leq f(x,k)$ for all $x \in \mathbb{R}^D$, and if the Radon-Nikodym derivative $\frac{d P_j}{d P_k}$ is bounded almost everywhere with respect to $P_k$ by a finite constant $C > 0$, then
\begin{equation}\label{eq:measure}
E[f(X, Z) | Z = j] \leq C E[f(X, Z) | Z = k].
\end{equation} 
\end{lemma}


If the conditional probability distribution of $X$ given $Z = j$ has density $p_{X | Z = j}(x)$, and $X$ given $Z = k$ has density $p_{X | Z = k}(x)$, then Lemma \ref{lem:measure} says that if the likelihood ratio $\frac{p_{X | Z = j}(x)}{p_{X | Z = k}(x)} \leq C$ almost everywhere with respect to $P_k$ for finite $C > 0$, then the one-sided statistical parity bound (\ref{eq:measure}) holds.

Fig. \ref{fig:theory} illustrates Lemma \ref{lem:measure} with an example: suppose $Z$ is categorical and denotes either a poorer or richer high school, and suppose $D = 1$ with $X \in [0,20]$ being the number of hours of extracurricular activities a student does each week. Let $f(X,Z)$ be a score used to determine a student's admission to some college. Then if $f$ is monotonic in $Z$ such that $f$ never gives a lower admissions score to a poorer student if their $X$ value is the same as a richer student, then Lemma 1 says that the average score for richer students will be no more than $C$ times the average score for poorer students, where $C$ is given by the maximum ratio of the two distributions over $X$.   
 


While Lemma 1 shows that monotonicity implies a bound on the one-sided statistical parity violations, the converse does not hold: a model satisfying statistical parity can have arbitrarily high monotonicity violations (proof in Appendix \ref{app:paritycounter}). This can be ethically problematic if overlooked by practitioners.

While Lemma \ref{lem:measure} provides a \emph{worst case bound} on the statistical parity violations of monotonic functions, we next ask, can imposing monotonicity ever make statistical parity violations worse? Lemma \ref{lem:moreunfair} shows that for any model $f$, the monotonic projection of $f$ cannot have worse statistical parity violations \textit{on average}.

\begin{lemma}\label{lem:moreunfair}
Let $f: \mathcal{X} \times \mathcal{Z} \to \mathbb{R}$, where $\mathcal{X} \subseteq \mathbb{R}^D$, $\mathcal{Z} \subseteq \mathbb{R}$. Assume that $\mathcal{X},\mathcal{Z}$ are both finite, with $X \in \mathcal{X}$, $Z \in \mathcal{Z}$. Let $\tilde{f}$  be the projection of $f$ onto the set of functions over $\mathcal{X} \times \mathcal{Z}$ that are monotonic with respect to $Z$ such that for $j \leq k$, $f(x,j) \leq f(x,k)$. For $z_{(i)} \in \mathcal{Z}$, let $z_{(1)} \leq z_{(2)} \leq ... \leq z_{(|\mathcal{Z}|)}$. Define the average statistical parity violation:
\begin{align*} R_f \stackrel{\triangle}{=} &\sum_{i = 1}^ {|\mathcal{Z}|} \frac{E[f(X, Z) | Z = z_{(i)}] - E[f(X, Z) | Z = z_{(i+1)}]}{|\mathcal{Z}|}
\end{align*}
Then $R_{\tilde{f}} \leq R_f$.
\end{lemma}

When $|\mathcal{Z}| = 2$, Lemma \ref{lem:moreunfair} also bounds the worst case violation of $\tilde{f}$. However, for $|\mathcal{Z}|  > 2$, there is no such worst case guarantee, and there may be pairs $j,k$ where $\tilde{f}$ has a worse one-sided statistical parity violation than $f$ (proof in Appendix \ref{app:moreunfaircounter}). 

\subsection{Bounds for Binary Classifiers}
In binary classification, an example has an associated true label $Y \in \{0,1\}$, and the model outputs a binary decision $\hat{Y} \in \{0,1\}$. By definition, a monotonicity shape constraint on $Z$ implies a one-sided bound on the conditional probabilities: for any $j \leq k$ and for all $x$, $
P(\hat{Y} = 1 | X = x, Z = j) \leq P(\hat{Y} = 1 |X = x, Z = k)$. 


Because the label $Y$ can be modeled as a Bernoulli random variable, the goal of statistical parity is equivalent to \emph{marginal independence}, that is, for any $j,k$, $P(\hat{Y} = 1 | Z = j) = P(\hat{Y} = 1 | Z = k)$. Correspondingly, the bound (\ref{eqn:measure}) on the statistical parity becomes a bound on the marginal probabilities: $P(\hat{Y} =1 | Z = j) \leq C P(\hat{Y}=1 | Z = k)$.

For binary classifiers, we can give a more explicit bound:

\begin{lemma}\label{lem:binary} Suppose $X$ is a continuous (or with a straightforward extension, discrete) random variable, and let $\mathcal{S}$ be a nonempty set such that for all $x \in \mathcal{S}$, the joint probability density values $p_{X, \hat{Y} | Z = z}(x,1) > 0$ for $z=j,k$.  Suppose we have monotonicity where $f(x,j) \leq f(x,k)$ for $j \leq k$ for all $x \in \mathcal{S}$. For a binary classifier this implies $P(\hat{Y} = 1 | X = x, Z = j) \leq P(\hat{Y} = 1 |X = x, Z = k)$. Then we can bound one-sided statistical parity as follows:
\begin{align*}\label{eq:binary}
    \frac{P(\hat{Y}=1 | Z = j)}{P(\hat{Y}=1 | Z = k) } \leq \inf_{x \in \mathcal{S}} \frac{p_{X|Z=j}(x) p_{X|\hat{Y}=1,Z=k}(x)}{p_{X|Z=k}(x)p_{X|\hat{Y}=1,Z=j}(x)}
\end{align*}
\end{lemma}


The bound in Lemma (\ref{lem:binary}) contains two likelihood ratios: $\frac{p_{X|Z=j}(x)}{p_{X|Z=k}(x)}$ and $\frac{p_{X|\hat{Y}=1,Z=k}(x)}{p_{X|\hat{Y}=1,Z=j}(x)}$. The first is the same as in Lemma $\ref{lem:measure}$. The second is the inverse of that likelihood ratio, conditioned on $\hat{Y} = 1$. When the first likelihood ratio is low, the second inverse likelihood ratio may be high, producing a trade-off between these two ratios. We describe an example in Appendix \ref{app:tradeoff}.

Similarly, for \textit{equal opportunity} \citep{hardt:2016}, we have Lemma \ref{lem:eqopp}:

\begin{lemma}\label{lem:eqopp} Let $Y \in \{0,1\}$ be a random variable representing the target. Let $\mathcal{S}$ be a nonempty set such that for all $x \in \mathcal{S}$, the following joint probability density values are non-zero for $z=j,k$: $p_{X,Y, \hat{Y} | Z = z}(x,1,1) > 0$ and $p_{X,Y| \hat{Y}=1, Z = z}(x,1) > 0$. Then,
\begin{align*}
    &\frac{P(\hat{Y}=1 | Y=1, Z = j)}{P(\hat{Y}=1 | Y=1, Z = k)} \leq \inf_{x \in \mathcal{S}} \frac{c_j(x)  }{c_k(x)} \\
    &\textrm{where } c_z(x) = \frac{p_{X|Z=z}(x) P(Y=1|\hat{Y}=1,Z=z)}{p_{X|\hat{Y}=1,Z=z}(x)P(Y=1|Z=z)}
\end{align*}
\end{lemma}


We supplement these bounds with empirical results.

\section{Experiments}\label{sec:experiments}
We demonstrate experimentally with three public datasets that \textit{(i)} nonlinear ML can violate common ethical policies or norms, \textit{(ii)}  training with monotonicity constraints can be used to impose such policies to the extent that the choice of model inputs $x$ enables,  \textit{(iii)} test accuracy may not be hurt,  \textit{(iii)} statistical fairness violations may be reduced. 

\textbf{Model training details:}
Code for these experiments and further tutorials for using the open-source TensorFlow Lattice 2.0 library \citep{TFLatticeBlogPost} are available at \url{https://github.com/tensorflow/lattice/blob/master/docs/tutorials/}. All experiments used a nonlinear generalized additive model (GAM) \citep{HastieTibshirani:90}, also called a \emph{calibrated linear} model in the TensorFlow Lattice, of the form $f(x) = \sum_{d=1}^D c_d(x[d]; \beta_d)$,
where each $c_d(\cdot)$ is an one-dimensional piecewise-linear function parameterized by $K$ interpolated (key, value) pairs where the keys are set to match the $K$ quantiles of the training data and their corresponding values  $\beta_d \in R^K$ are trained \citep{GuptaEtAl:2016}. We used a fixed $K=20$ parameters for each of the $D$ one-dimensional transforms, for all experiments.  All $DK$ parameters were jointly trained using projected stochastic gradient descent. For each input $d$, one can choose to constrain $f(x)$ to be monotonically increasing with respect to $d$ by constraining its one-dimensional curve $c_d(x[d])$ to be monotonic \citep{GuptaEtAl:2016}. The TensorFlow Lattice package enables fitting more flexible monotonic models, but we felt it was most compelling to show that these monotonicity violations occur even with simple nonlinear models like GAMs, which are popular in the statistics literature. We randomly uniformly split each dataset into 70\% training, 10\% validation, and 20\% test examples.  We validated the learning rates using grid search by powers of 10 ensuring that the optimal learning rates did not fall on the extremes, and trained for 1000 epochs (more than sufficient for convergence). 


\subsection{Law School Experiments}\label{sec:Law}
In Section 1 we partially described an experiment using the \textit{Law School Admissions} dataset \citep{Wightman:1998}. The dataset has 27,234 total law students, and we use only two features in all models: GPA and LSAT scores. We plot the densities of these conditioned on whether the student passed the bar or not in Fig. \ref{fig:law_data} in Appendix \ref{app:law}.  Models were trained as described above, where for the constrained model we constrained both LSAT score and undergraduate GPA to only be positive evidence, \emph{ceterus paribus}. The two-layer neural network and gradient boosted trees models were trained with a similar number of model parameters to the GAMs and no monotonicity constraints. Resulting models were shown in Fig. \ref{fig:law_models}, and train/test accuracies in Table \ref{tab:law}.

\subsection{Credit Default Experiments}\label{sec:creditExperiments}
Next, we consider the Default of Credit Card Clients benchmark dataset from the UCI repository \citep{Lichman:2013, Yeh:2009}. The data was collected from 30,000 Taiwanese credit card users and contains a binary label of whether or not a user defaulted on a payment in a time window.  Features include marital status, gender, education, and how long a user is behind on payment of their existing bills, for each of the months of April-September 2005.

Repayment status is the integer number of months it has been since the user has repaid, and negative values mean the user has already repaid. Here we illustrate using monotonicity constraints to avoid \emph{unfair penalization}: if the model were to be used to determine a user's credit score, it could feel unfair to many if they were penalized for paying their bills sooner, all else equal. Thus, we apply a monotonicity constraint that keeps the model from penalizing early payments. 

Figure \ref{fig:trainingdata} \emph{(top)} shows the average and standard deviation of the default rate of the training examples as a function of the months since the bills were paid. The data is very noisy for $3^+$ months overdue payment (there are only 122 such training examples), and these noisy averages do not follow the reasonable principle of predicting a higher default rate if your bills are more over-due. 

For ease of visualization, our first experiment uses only $D=2$ features: marital status and April repayment status, with a monotonicity shape constraint on the repayment status to ensure that paying your bills on time doesn't hurt you.  Our second experiment uses all $D=24$ features, and we impose monotonicity shape constraints on all 6 repayment features to guarantee that the person is not penalized for paying early/on-time during any of the 6 months.

Fig. \ref{fig:creditmodels} \emph{(top)} shows the predicted default rate for the unconstrained GAM. It mimics the average training labels, and thus unfairly rewards people who are 5-6 months overdue on their bills with a lower defaulting score than people who are only 2-3 months overdue. However, Figure \ref{fig:creditmodels} \emph{(bottom)} shows the GAM trained with a monotonicity shape constraint: as requested, it does not penalize people for paying their bills early.

Table \ref{tab:credit} shows that adding the monotoniciy constraints for the Credit Default problem had only a tiny effect on the train and test accuracy for the $D=2$ experiments. For the $D=24$ experiment, the test accuracy is slightly worse, but since the train accuracy was not hurt, we believe the lower test accuracy is simply due to the randomness of the sample.  


\subsection{Funding Proposals Experiments} \label{sec:fundingExperiments}
\begin{figure}[ht!]
\centering
\begin{tabular}{c}
Credit Default: Training Examples \\
\includegraphics[width=0.362\textwidth]{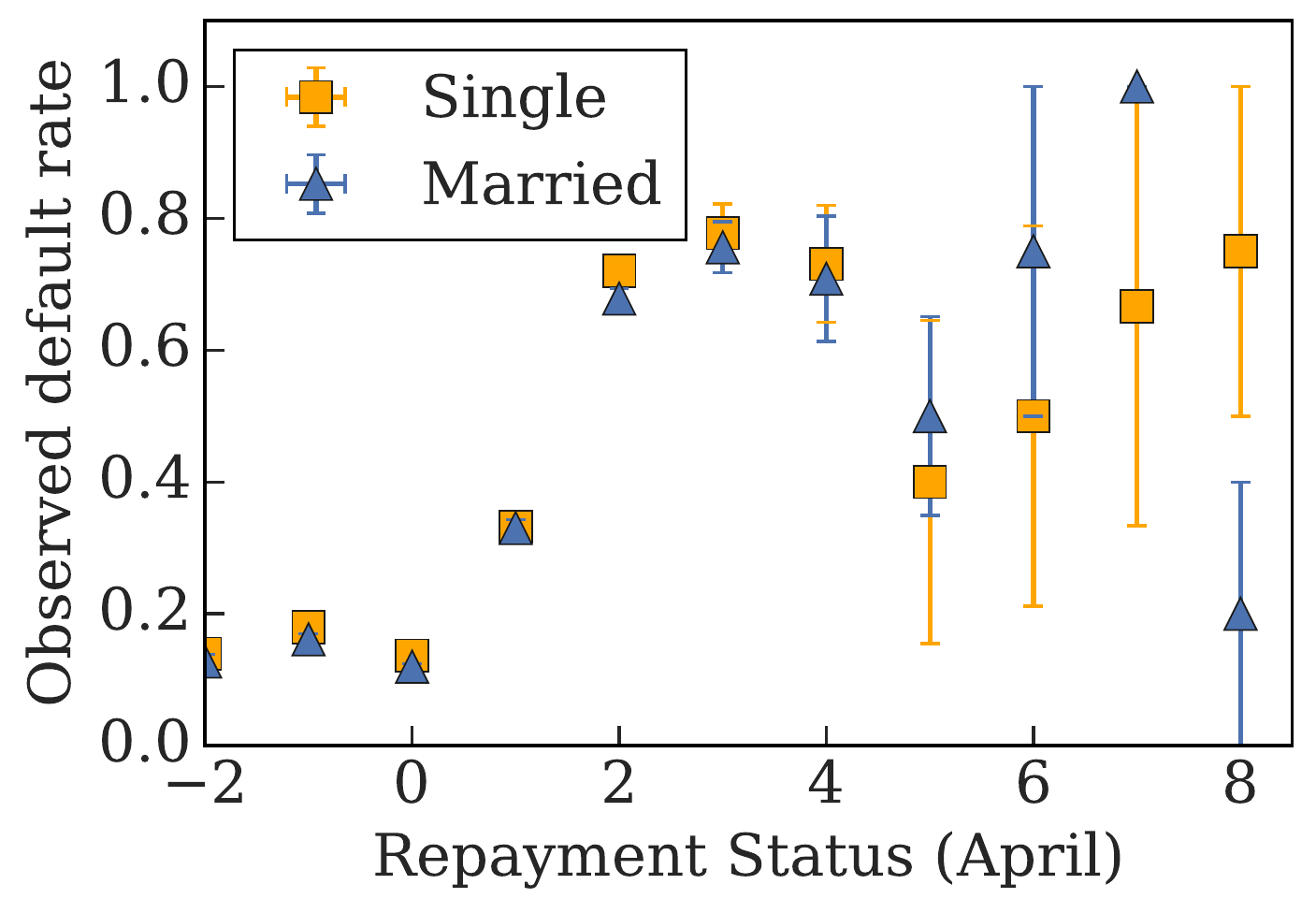} \\
Funding Proposals: Training Examples \\
\includegraphics[width=0.362\textwidth]{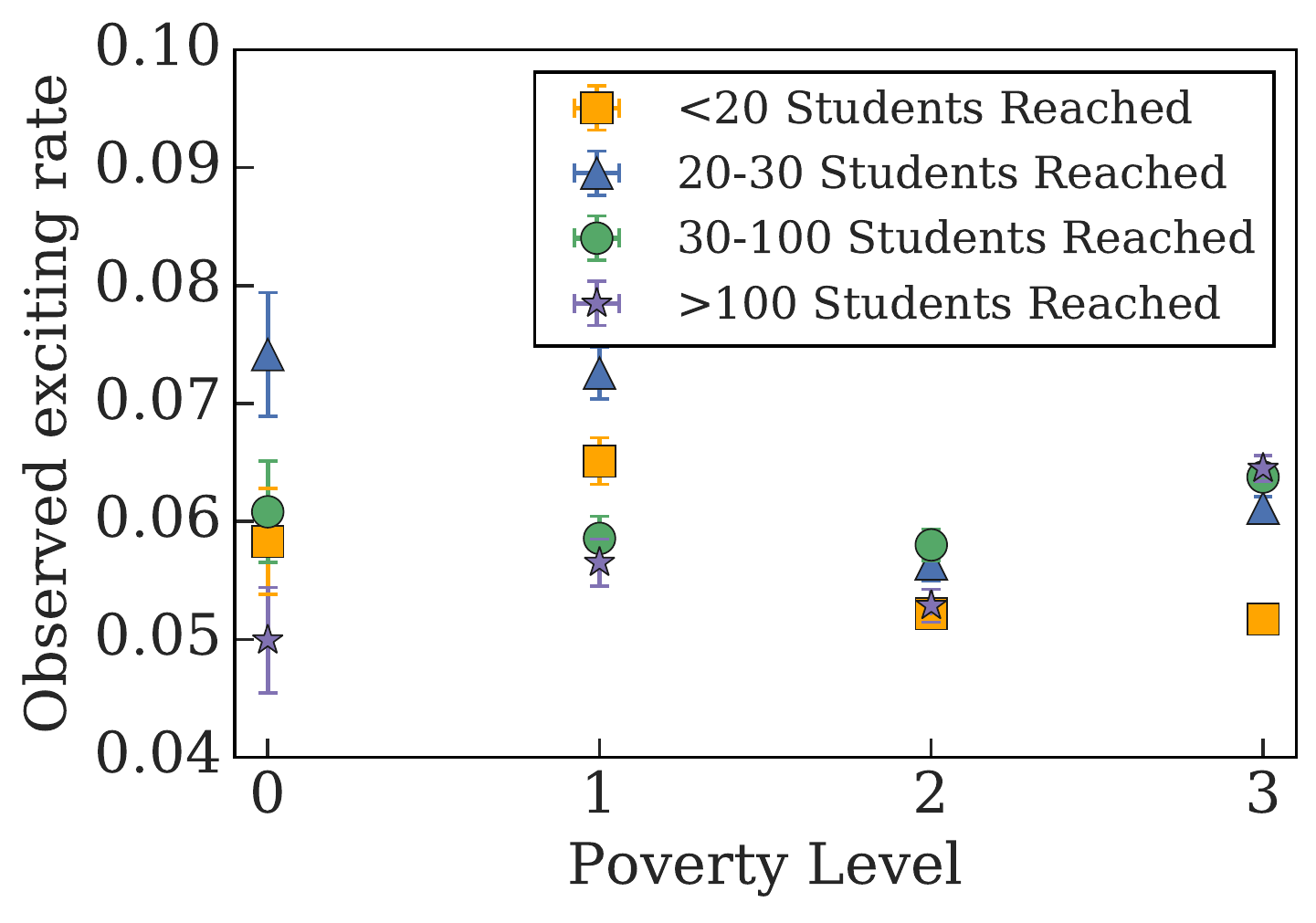} \\
\end{tabular}
\caption{Mean and standard error of the label as a function of the inputs over the training dataset.}
\label{fig:trainingdata} 
\end{figure}

\begin{figure}[ht!]
\centering
\begin{tabular}{c}
 Credit Default: Unconstrained Model Predictions \\ 
\includegraphics[width=0.362\textwidth]{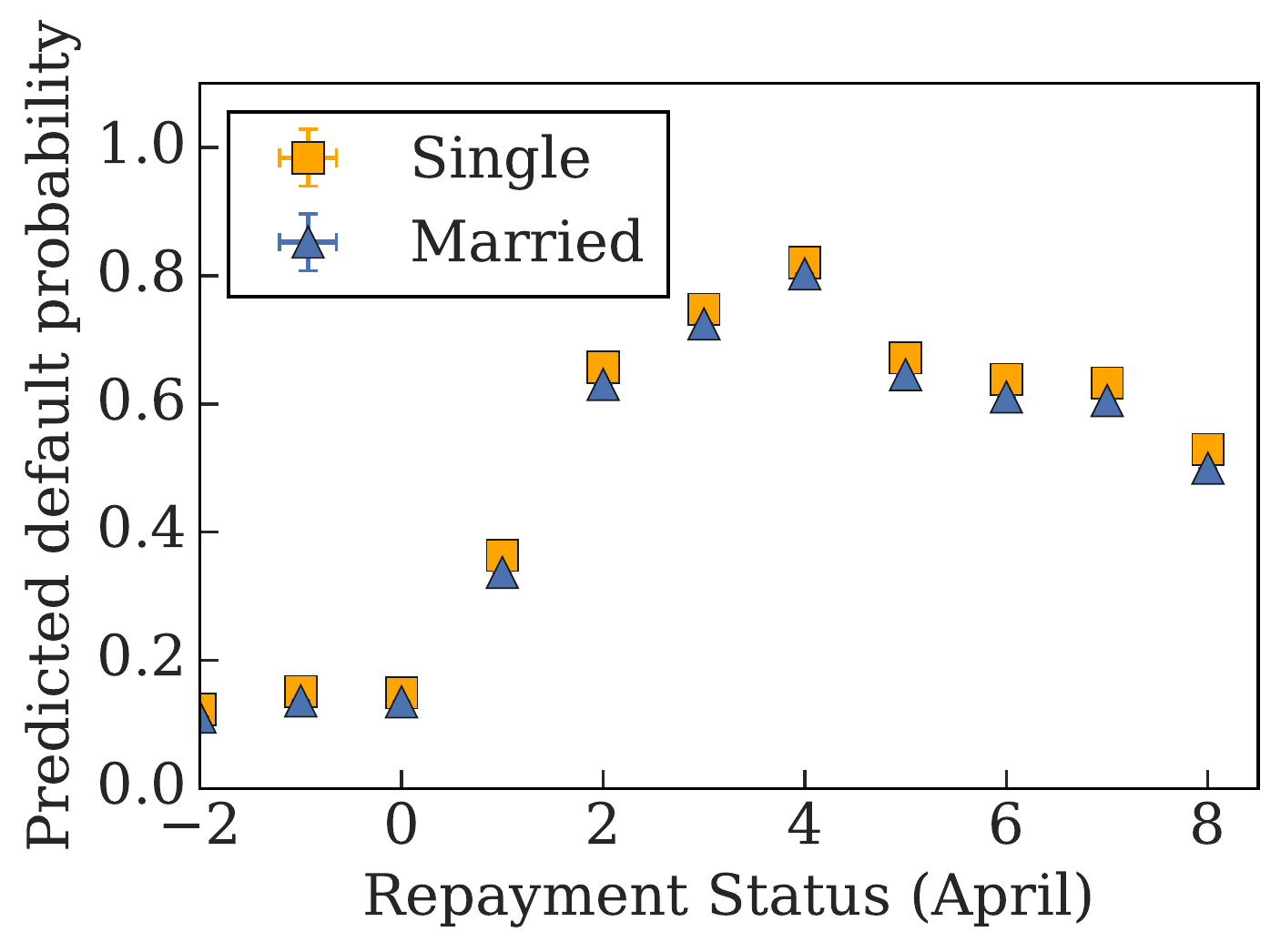} \\
Credit Default: Monotonic Model Predictions \\ \includegraphics[width=0.362\textwidth]{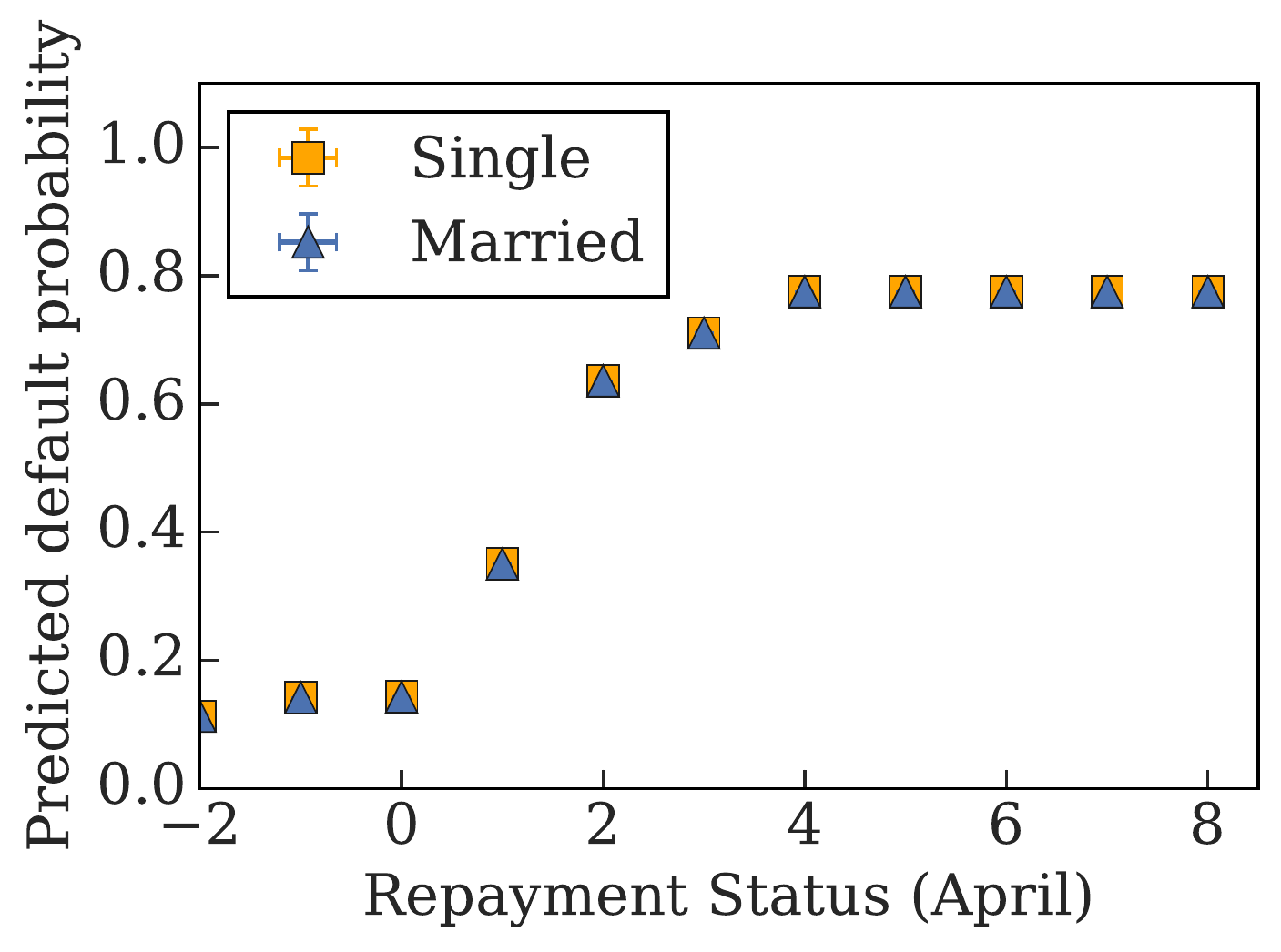} \\
\end{tabular}
\caption{Credit Default: unconstrained model predictions \textit{(top)} and constrained model predictions \textit{(bottom)}. }
\label{fig:creditmodels} 
\end{figure}

Next we demonstrate the use of monotonicity constraints to favor the less fortunate. We use the dataset from the KDD Cup 2014: Predicting Excitement at DonorsChoose.org \citep{KDD:2014}. DonorsChoose.org is a platform through which teachers in K-12 schools can request funding and materials for proposed projects. Donors can search for and donate to projects. The dataset contains 619,327 examples of projects proposed by teachers. The label is a binary label that represents the outcome of the project, where the positive class means the project was deemed ``exciting'', a definition determined by DonorsChoose.org that includes whether the project was fully funded. Only $5.91\%$ of examples are labeled ``exciting''. Figure \ref{fig:trainingdata} \emph{(bottom)} shows the mean and std. dev. of the training examples for 2 of the 28 features: poverty level and number of students impacted. A machine learning model trained on this dataset could be used to rank projects to display to potential donors. Standard machine learning trained on just these two features would not favor schools at higher poverty levels, or projects with greater impact.  We compare to training with monotonicity constraints for those two principles, both on just the shown two features, and on all 28 features (but only these two features constrained).

The results in Table \ref{tab:donors} for the AUC metric (the same metric used in the KDD Cup 2014 competition) show that adding monotonicity constraints to prefer poorer schools and greater student impact did not hurt the test AUC. Figures comparing the standard and constrained $D=2$ models are in Appendix \ref{app:donors}. 

\begin{table}[!ht]
\centering
\begin{tabular}{ll|ll}
\hline
\# features & \# shape  & Train  & Test\\
& constraints & Acc. & Acc. \\
\hline
$2$& 0  & $82.07\%$  & $81.55\%$\\
$2$& 1  & $82.06\%$  & $81.60\%$ \\
$24$& 0  & $82.44\%$  & $82.02\%$\\
$24$& 6 & $82.35\%$  & $80.86\%$\\
\hline
\end{tabular}
\caption{Credit Default experiment results, where the monotonicity shape constraints ensure the model does not penalize people for paying their bills earlier.} 
\label{tab:credit}
\end{table}



\begin{table}[!hb]
\centering
\begin{tabular}{ll|ll}
\hline
\# features & \# shape  & Train  & Test\\
& constraints & AUC  & AUC \\
\hline
$2$ & 0 & $0.520$ & $0.517$\\
$2$& 2  & $0.514$  & $0.518$ \\
$28$& 0 & $0.752$  & $0.746$\\
$28$& 2  & $0.751$  & $0.746$ \\ 
\hline
\end{tabular}
\caption{Funding Proposals experiment results, where the monotonicity shape constraints ensure that the model gives higher scores to projects for higher poverty schools or impacting more students, all else equal.} 
\label{tab:donors}
\end{table}

\subsubsection{Comparison to Statistical Fairness}
In Section 6 we gave \emph{theoretical} results on how monotonicity constraints can bound and affect \textit{one-sided statistical parity} and \textit{one-sided equal opportunity} fairness violations. In Table \ref{tab:funding} we show empirically what happens to the one-sided (1-s) statistical parity and equal opportunity violations (defined below) for the Funding Proposals experiment, where the protected groups $j,k \in \{0,1,2,3\}$ are four different \textit{poverty levels}, which we treat as four different ordinal protected groups for calculating these one-sided statistical metrics.

Max \textit{One-sided Statistical Parity} Violation:
\hspace{10mm}
\begin{equation}\label{eq:maxstatpar}
    \displaystyle\max_{j<k} (0, P(\hat{Y}=1 | Z = j) - P(\hat{Y}=1 |  Z = k))
\end{equation}
Max \textit{One-sided Equal Opportunity} Violation: 
\hspace{10mm} 
\begin{align}
\begin{split}
\displaystyle\max_{j<k} (0, \; &P(\hat{Y}=1 |Z = j, Y=1) \\
&- P(\hat{Y}=1 | Z = k, Y=1))
\end{split}\label{eq:maxeqopp}
\end{align}

Table \ref{tab:funding} shows that the monotonically constrained models do lower violations of both fairness goals. The improvement is smaller when the model has $D=28$ features than $D=2$ features, which we believe is due to a weakening of the maximum likelihood ratios as described by Lemmas \ref{lem:binary} and \ref{lem:eqopp}.


\begin{table}[h!]
\centering
\begin{tabular}{llll}
    \hline
$D$ &  \# shape  & Max 1-s Stat.   & Max 1-s Eq.   \\
 &  constraints  &   Par. Viol. (\ref{eq:maxstatpar}) & Opp. Viol. (\ref{eq:maxeqopp}) \\
    \hline
 $2$ & 0 & 0.00704 & 0.00707  \\
    $2$ & 2 & 0.00017 & 0.00037 \\
   $28$ & 0 & 0.00751 & 0.00397 \\
$28$ & 2 & 0.00261 & 0.00331 \\
    \hline
\end{tabular}
\caption{Statistical Fairness Violations for Funding Proposals Experiment.}
\label{tab:funding}
\end{table}

\section{Discussion} \label{sec:conclusions}
We have demonstrated that nonlinear machine learned models can easily overfit noise or learn bias in a way that violates social norms or ethics about whether certain inputs should be allowed to negatively affect a score or decision. We have also shown that this problem can be ameliorated by training with monotonicity constraints to reflect the desired principle. 

An advantage of monotonicity constraints is that their effect does not depend on the data distribution, so there are no questions of generalization to test data as there are with statistical fairness measures. While enforcing such constraints addresses deontological rather than consequentialist ethics, we have also shown theoretically and experimentally that monotonicity constraints can can improve or bound (consequentialist) one-sided statistical fairness violations. 


Monotonicity constraints are fairly easy to explain and reason about for laypeople, as illustrated in the examples in this paper.  The examples also illustrate the broad applicability of constraints to situations where the model should avoid unfair penalization of good attributes, and favor the less fortunate.


We conclude that monotonicity constraints are a necessary and useful tool for creating responsible AI, but certainly not sufficient or applicable to all situations. For this method to be completely effective, all relevant features must be identified and constrained.  Non-sensitive information can be highly correlated with sensitive information, causing indirect discrimination \citep{Hajian:2013}, which is also a problem for statistical fairness measures \citep{hardt:2016}. This method is also not directly applicable to unordered inputs like addresses, photos, or voice signals. Thus, this will be one of many tools and strategies needed to achieve responsible AI.  



%


\clearpage

    %
    %
    %
    %
    %
    %

\bibliography{references}

\clearpage
\newpage
\appendix

\section{Proofs}\label{app:proofs}

\begin{replemma}{lem:measure} 
Let $(\Omega, \mathcal{F})$ be a measurable space with a regular conditional probability property, and let $X: \Omega \to \mathbb{R}^D$, $Z: \Omega \to \mathbb{R}$ be $\mathcal{F}$-measurable random variables. Suppose $P_j$ and $P_k$ are $\sigma$-finite probability measures on $(\Omega, \mathcal{F})$, where $P_j$ denotes the conditional probability measure of $X$ given that $Z = j$, and $P_k$ denote the same for $Z = k$, and $P_j$ is absolutely continuous with respect to $P_k$. Let $f: \mathbb{R}^D \times \mathbb{R} \to \mathbb{R}$ be defined as in Section \ref{sec:monotonic}, and $f(x,z) \geq 0$ for all $x \in \mathbb{R}^D$, $z \in \mathbb{R}$. If the function $f$ satisfies monotonicity in the second argument such that $f(x,j) \leq f(x,k)$ for all $x \in \mathbb{R}^D$ and for $j \leq k$, and if the Radon Nikodym derivative $\frac{d P_j}{d P_k}$ is bounded almost everywhere with respect to $P_k$ by a finite constant $C > 0$, then
\begin{equation}\label{eqn:measure}
E[f(X, Z) | Z = j] \leq C E[f(X, Z) | Z = k].
\end{equation} 
\end{replemma}

\begin{proof}
Under Lemma \ref{lem:measure}'s assumptions, 
\begin{align*}
E[f(X,Z) | Z = j] &= \int_{\mathbb{R}^D} f(x,j)  dP_j \\
    &\leq \int_{\mathbb{R}^D} f(x,k) dP_j \\
    &= \int_{\mathbb{R}^D} f(x,k) \frac{dP_j}{dP_k }dP_k \\
    &\leq C\int_{\mathbb{R}^D} f(x,k) dP_k \\
    &= C E[f(X,Z) | Z = k]. \hspace{40mm} 
\end{align*}
The second inequality follows from monotonicity, and the third by the Radon Nikodym theorem since $P_j << P_k$. 
\end{proof}

\begin{replemma}{lem:moreunfair} 
Let $f: \mathcal{X} \times \mathcal{Z} \to \mathbb{R}$, where $\mathcal{X} \subseteq \mathbb{R}^D$, $\mathcal{Z} \subseteq \mathbb{R}$. Assume that $\mathcal{X},\mathcal{Z}$ are both finite, with $X \in \mathcal{X}$, $Z \in \mathcal{Z}$. Let $\tilde{f}$  be the projection of $f$ onto the set of functions over $\mathcal{X} \times \mathcal{Z}$ that are monotonic with respect to $Z$ such that for $j \leq k$, $f(x,j) \leq f(x,k)$. For $z_{(i)} \in \mathcal{Z}$, let $z_{(1)} \leq z_{(2)} \leq ... \leq z_{(|\mathcal{Z}|)}$. Define the average \textit{statistical parity} violation:
\begin{align*} R_f \stackrel{\triangle}{=} &\sum_{i = 1}^ {|\mathcal{Z}|} \frac{E[f(X, Z) | Z = z_{(i)}] - E[f(X, Z) | Z = z_{(i+1)}]}{|\mathcal{Z}|}
\end{align*}
Then $R_{\tilde{f}} \leq R_f$.
\end{replemma}

\begin{proof}
Let $\tilde{f}: \mathcal{X} \times \mathcal{Z} \to \mathbb{R}$ be the projection of $f$ onto the class of functions monotonic in the second argument, defined as follows:
\begin{align}
\begin{split}
\tilde{f} =\; &\underset{f'}{\arg\min} \; ||f - f'||\\
&\text{s.t.}\; f'(x,j) \leq f'(x,k) \; \forall j,k \in \mathcal{Z};  j \leq k
\end{split}\label{eq:projection}
\end{align}
where $$||f - f'||^2 = \sum_{x \in \mathcal{X}, z \in \mathcal{Z}} (f(x,z) - f'(x,z) )^2 .$$

The projection $\tilde{f}$ can be computed in $O(|\mathcal{X}| |\mathcal{Z}|)$ time using the pool-adjacent-violators algorithm from isotonic regression \citep{Ayer:1955, Kruskal:1964}, since a one dimensional projection can be done independently in  $O(|\mathcal{Z}|)$ time for each $x \in \mathcal{X}$.

$R_f$ is a telescoping sum: 
\begin{align*}
    R_f &= \frac{E[f(X, Z) | Z = z_{(1)}] - E[f(X, Z) | Z = z_{(|\mathcal{Z}|)}]}{|\mathcal{Z}|}
\end{align*}
For discrete $X$ and $Z$, we have $$E[f(X, Z) | Z = j] = \sum_{x \in \mathcal{X}} f(x, j) P(X = x | Z = j)$$
which implies
\begin{align*}
   R_f = \frac{1}{|\mathcal{Z}|} \sum_{x \in \mathcal{X}} \bigg( &f(x, z_{(1)}) P(X = x | Z = z_{(1)}) \\
   - &f(x, z_{(|\mathcal{Z}|)}) P(X = x | Z = z_{(|\mathcal{Z}|)}) \bigg).
\end{align*}

We now show that $\tilde{f}(x, z_{(1)}) \leq f(x, z_{(1)})$, and $\tilde{f}(x, z_{(|\mathcal{Z}|)}) \geq f(x, z_{(|\mathcal{Z}|)})$: \\ Suppose $\tilde{f}(x, z_{(1)}) > f(x, z_{(1)})$. Then we can set $\tilde{f}'(x,z_{(1)}) = f(x,z_{(1)})$ without violating the monotonicity constraints, and $||f - \tilde{f}' || < || f - \tilde{f}||$, which contradicts that $\tilde{f}$ solves (\ref{eq:projection}). A similar argument can be made for $z_{(|\mathcal{Z}|)}$.

Since $\tilde{f}(x, z_{(1)}) \leq f(x, z_{(1)})$ and \\$\tilde{f}(x, z_{(|\mathcal{Z}|)}) \geq f(x, z_{(|\mathcal{Z}|)})$, we have 
\begin{align*}
    f(x,  z_{(1)}) & P(X = x | Z = z_{(1)}) \\
     &-f(x, z_{(|\mathcal{Z}|)}) P(X = x | Z = z_{(|\mathcal{Z}|)}) \\
    \geq \tilde{f}(x, z_{(1)}) & P(X = x | Z = z_{(1)}) \\
     &- \tilde{f}(x, z_{(|\mathcal{Z}|)}) P(X = x | Z = z_{(|\mathcal{Z}|)}) 
\end{align*}

Since the above inequality is true for all $x$, it holds for the sum over $x \in \mathcal{X}$, therefore $R_{\tilde{f}} \leq R_f$.
\end{proof}

\begin{replemma}{lem:binary} Suppose $X$ is a continuous (or with a straightforward extension, discrete) random variable, and let $\mathcal{S}$ be a nonempty set such that for all $x \in \mathcal{S}$, the joint probability density values $p_{X, \hat{Y} | Z = z}(x,1) > 0$ for $z=j,k$.  Suppose we have monotonicity where $f(x,j) \leq f(x,k)$ for $j \leq k$ for all $x \in \mathcal{S}$. For a binary classifier this implies $P(\hat{Y} = 1 | X = x, Z = j) \leq P(\hat{Y} = 1 |X = x, Z = k)$. Then we can bound \textit{one-sided statistical parity} as follows:
\begin{align*}\label{eq:binary}
    \frac{P(\hat{Y}=1 | Z = j)}{P(\hat{Y}=1 | Z = k) } \leq \inf_{x \in \mathcal{S}} \frac{p_{X|Z=j}(x) p_{X|\hat{Y}=1,Z=k}(x)}{p_{X|Z=k}(x)p_{X|\hat{Y}=1,Z=j}(x)}
\end{align*}
\end{replemma}

\begin{proof}
Fix $x \in \mathcal{S}$. By Bayes' theorem and monotonicity,
\begin{align*}
    P(\hat{Y}&=1 | Z = j) \\
    &= P(\hat{Y}=1 | X = x, Z = j) \frac{p_{X|Z=j}(x)}{p_{X|\hat{Y}=1,Z=j}(x)}  \\
    &\leq P(\hat{Y}=1 | X = x, Z = k) \frac{ p_{X|Z=j}(x)}{p_{X|\hat{Y}=1,Z=j}(x)}  \\
    &= P(\hat{Y}=1 | Z = k) \frac{p_{X|\hat{Y}=1,Z=k}(x)}{p_{X|Z=k}(x)} \frac{p_{X|Z=j}(x)}{p_{X|\hat{Y}=1,Z=j}(x)} \\
\end{align*}
Since the inequality holds for all $x \in \mathcal{S}$, the tightest bound holds for the infimum.
\end{proof}

\begin{replemma}{lem:eqopp} Let $Y \in \{0,1\}$ be a random variable representing the target. Let $\mathcal{S}$ be a nonempty set such that for all $x \in \mathcal{S}$, the following joint probability density values are non-zero for $z=j,k$: $p_{X,Y, \hat{Y} | Z = z}(x,1,1) > 0$ and $p_{X,Y| \hat{Y}=1, Z = z}(x,1) > 0$. Then,
\begin{align*}
    &\frac{P(\hat{Y}=1 | Y=1, Z = j)}{P(\hat{Y}=1 | Y=1, Z = k)} \leq \inf_{x \in \mathcal{S}} \frac{c_j(x)  }{c_k(x)} \\
    &\textrm{where } c_z(x) = \frac{p_{X|Z=z}(x) P(Y=1|\hat{Y}=1,Z=z)}{p_{X|\hat{Y}=1,Z=z}(x)P(Y=1|Z=z)}
\end{align*}
\end{replemma}

\begin{proof} 

Let $\mathcal{S}$ be a nonempty set such that for all $x \in \mathcal{S}$, the following joint probability density values are non-zero for $z=j,k$: 

$p_{X,Y, \hat{Y} | Z = z}(x,1,1) > 0$ and $p_{X,Y| \hat{Y}=1, Z = z}(x,1) > 0$

Fix $x \in \mathcal{S}$. 

Suppose we have a monotonic binary classifier, where $P(\hat{Y} = 1 | X=x, Z=j) \leq  P(\hat{Y} = 1 | X=x, Z=k)$ for $j \leq k$.

By Bayes' theorem, we have \\
\scalebox{0.8}{ $P(Y=1|Z=j)P(\hat{Y} = 1 | Y=1, Z=j) p_{X|Y=1,\hat{Y}=1,Z=j}(x) $} \\
\scalebox{0.8}{$= p_{X|Z=j}(x)P(\hat{Y} = 1 | X=x, Z=j)P(Y=1|X=x, \hat{Y} = 1, Z=j)$}

and
$p_{X|\hat{Y}=1,Z=j}(x) P(Y=1|X=x, \hat{Y}=1, Z=j)$ \\ 
$= p_{X|Y=1,\hat{Y}=1,Z=j}(x) P(Y=1|\hat{Y}=1,Z=j)$ 

Let $c_z(x) = \frac{p_{X|Z=z}(x) P(Y=1|\hat{Y}=1,Z=z)}{p_{X|Y=1,Z=z}(x)P(Y=1|Z=z)}$. This is well defined for $x \in \mathcal{S}$.

Combining both applications of Bayes' theorem and the monotonicity assumption:

\begin{align*}
    P(\hat{Y} &= 1 &&| Y=1, Z=j) \\
    &=&&\frac{p_{X|Z=j}(x) P(Y=1|X=x, \hat{Y} = 1, Z=j)}{P(Y=1|Z=j)p_{X|Y=1,\hat{Y}=1,Z=j}(x)}\\
    &\text{ }&&*P(\hat{Y} = 1 | X=x, Z=j) \\
    &=&&\frac{p_{X|Z=j}(x) P(Y=1|\hat{Y}=1,Z=j)}{P(Y=1|Z=j)p_{X|\hat{Y}=1,Z=j}(x)}\\
    &\text{ }&&*P(\hat{Y} = 1 | X=x, Z=j) \\
    &= &&c_j(x) P(\hat{Y} = 1 | X=x, Z=j) \\
    &\leq &&c_j(x) P(\hat{Y} = 1 | X=x, Z=k) \\
    &= &&\frac{c_j(x)}{c_k(x)} P(\hat{Y} = 1 | Y=1, Z=k) 
\end{align*}

Since this holds for all $x \in \mathcal{S}$, it holds for the infimum. \end{proof}

\section{Counterexamples} \label{sec:counterexamples}

To supplement Section \ref{sec:theory}, we give various counterexamples showing that certain relations between \textit{statistical parity} and monotonicity do not hold.

\subsection{Monotonicity does not imply statistical parity.}\label{app:moncounter}

\begin{figure}[h!]
\includegraphics[width=0.45\textwidth]{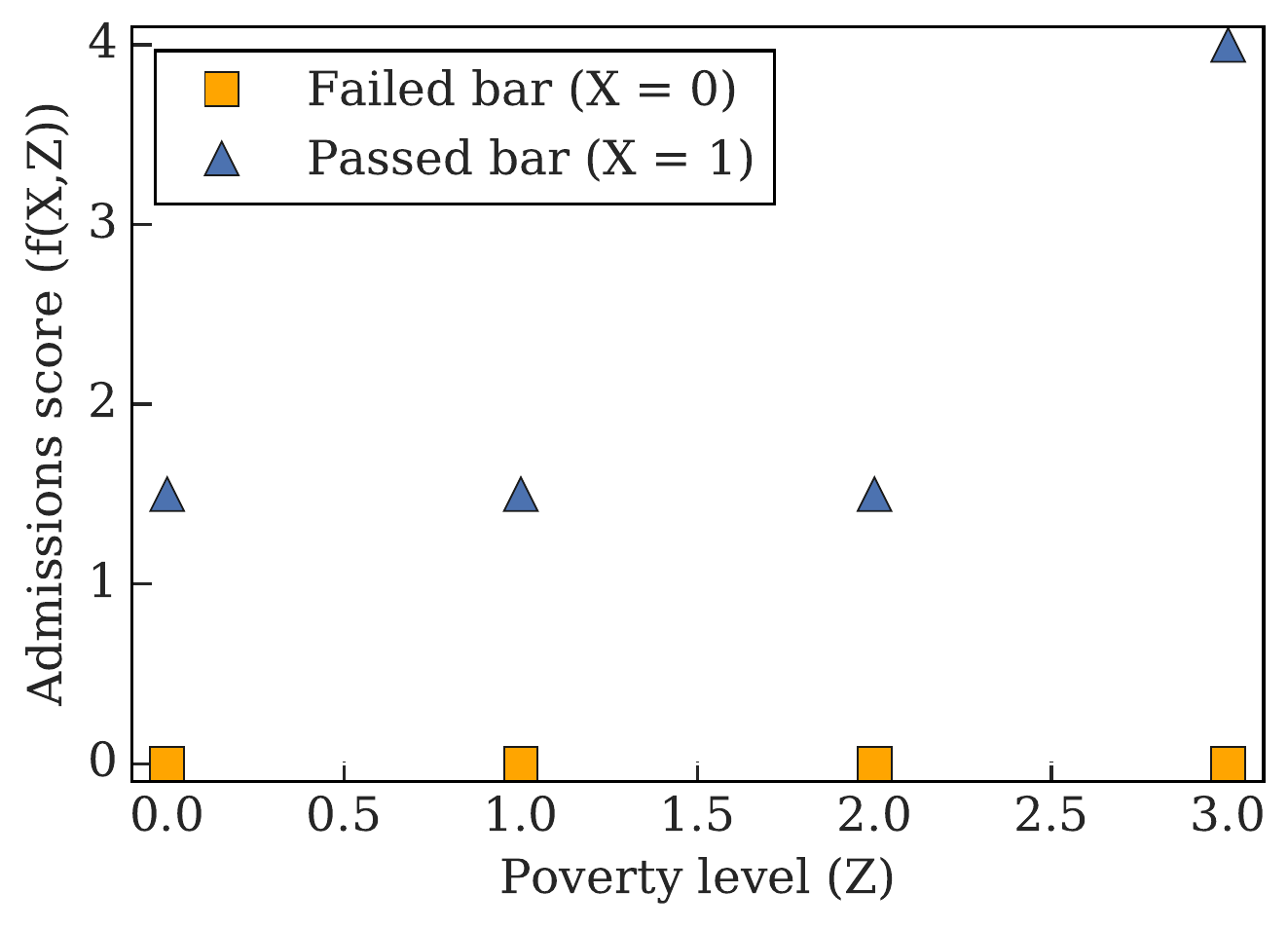}
\caption{Monotonic admissions scores for Counterexamples \ref{app:moncounter} and \ref{app:moreunfaircounter}.}
\label{fig:moncounter} 
\end{figure}

\begin{figure}[h!]
\begin{tabular}{c}
$P(X = x | Z = z)$ \\
\includegraphics[width=0.45\textwidth]{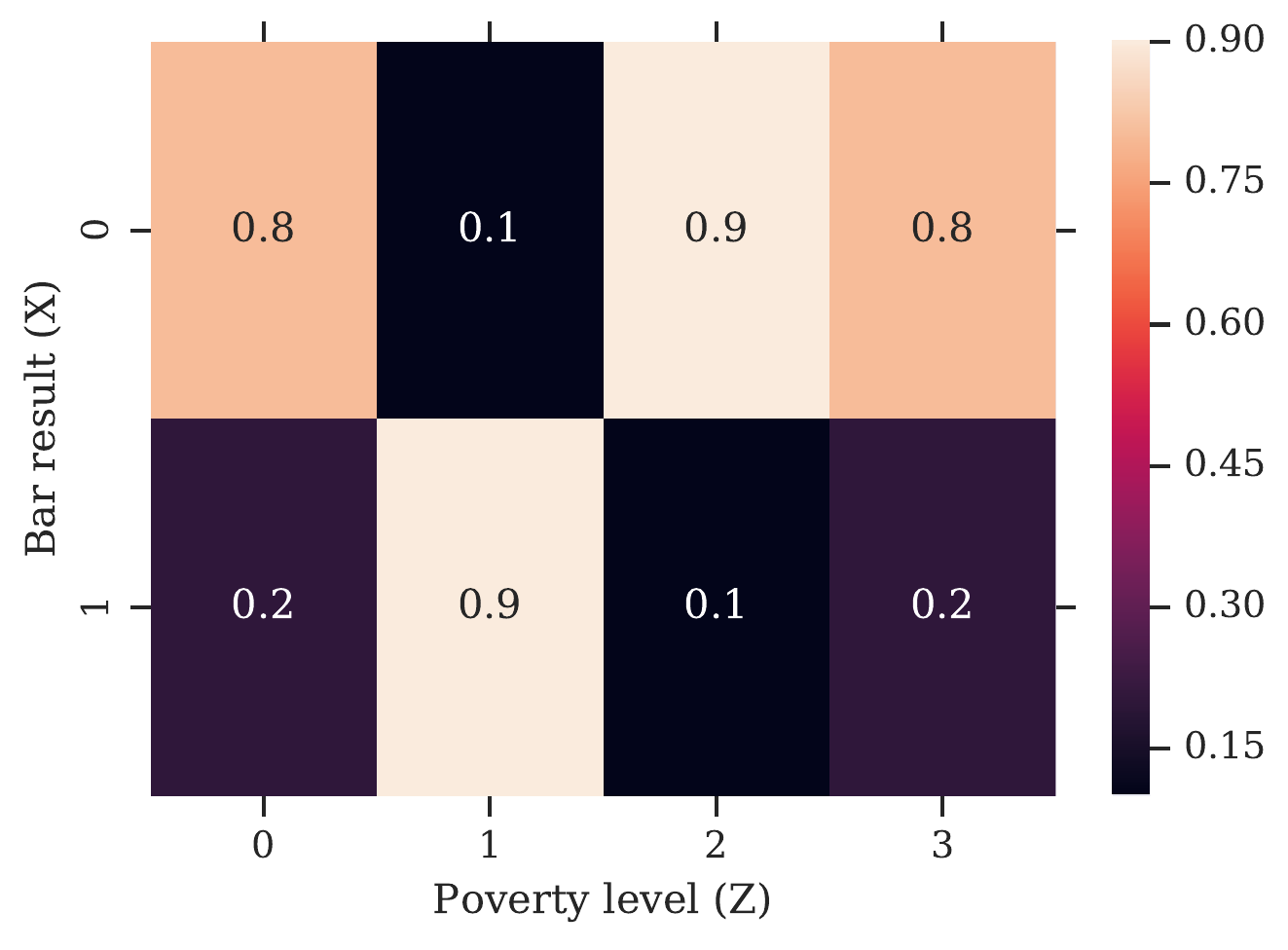} \end{tabular}
\caption{Distribution of $X,Z$ for Counterexamples \ref{app:moncounter} and \ref{app:moreunfaircounter}. The displayed values are $P(X = x | Z = z)$ for $X \in \{0,1\}$ and $Z \in \{0,1,2,3\}$.}
\label{fig:distcounter} 
\end{figure}

We show that monotonic function $f$ may violate \textit{one-sided statistical parity} by an example that illustrates Simpson's paradox. Suppose $X \in \{0,1\}$, where $X = 1$ means a law student passed the bar and $X = 0$ means the student did not. Let $Z \in \{0,1,2,3\}$ be the poverty level of the student, where $Z = 3$ represents the highest poverty level. Suppose $f(X,Z)$, or the admissions score, is monotonic in $Z$ and takes the values shown in Fig. \ref{fig:moncounter}. Suppose that the distributions $P(X = x | Z = z)$ are given by figure \ref{fig:distcounter}. Then the maximum \textit{one-sided statistical parity} violation is 
\begin{align*}
    &E[f(X,Z) | Z = 1] - E[f(X,Z) | Z = 2] \\
    &= f(0, 1)P(X = 0 | Z = 1) - f(0,2)P(X = 0 | Z = 2) \\
    &\;\;- f(1, 1)P(X = 0 | Z = 1) + f(1,2)P(X = 0 | Z = 2) \\
    &= 1.5(0.9) - 1.5(0.1) \\
    &= 1.2.
\end{align*}
Thus, there is a positive one-sided satistical parity violation even though $f(X,Z)$ is monotonic in $Z$. This violation comes from the fact that even though $f(0,1) \leq f(0,2)$, this is outweighed by the fact that $P(X = 0 | Z = 1) \geq P(X = 1 | Z = 2)$. This illustrates that for a monotonic function, the \textit{statistical parity} violation depends on the conditional probabilities $P(X = x | Z = z)$, and indeed Lemma \ref{lem:measure} bounds the one-sided statistcal parity violation by a ratio of conditional probabilities.

\subsection{Statistical Parity does not imply a bound on monotonicity violations.}\label{app:paritycounter}

We show that the converse of Lemma \ref{lem:measure} does not hold: a model that satisfies \textit{statistical parity} may have arbitrarily high monotonicity violations regardless of the likeihood ratio $C$. Suppose the distribution of men and women for a given height $x$ is equal for all heights, such that $C = 1$. Suppose that \textit{statistical parity} is satisfied such that men and women were equally likely to be selected for a sports team on average.\textit{Statistical parity} could hold if  the model accepted all men over some height $h$ that splits the population in half (say $h = 5'8''$), and accepted all women under height $h$. But then for a height less than $h$, $P(\hat{Y} = 1 | Z = \textrm{female}) = 0$ while $P(\hat{Y} = 1 | Z = \textrm{male}) = 1$, and for height over $h$,  $P(\hat{Y} = 1 | Z = \textrm{male}) = 0$ while $P(\hat{Y} = 1 | G = \textrm{female}) = 1$. Therefore, neither a positive nor a negative monotonicity constraint holds: there is no constant $C' > 0$ such that
$P(\hat{Y} = 1 |  X = x, Z = \textrm{male}) \leq C'P(\hat{Y} = 1 | X = x, Z = \textrm{female})$
or $P(\hat{Y} = 1 | X = x, Z = \textrm{female}) \geq C'P(\hat{Y} = 1 | X = x, Z = \textrm{male})$ for all $x$.

\subsection{Monotonic projection can be more unfair in the worst case.}\label{app:moreunfaircounter}
\begin{figure}[h!]
\includegraphics[width=0.45\textwidth]{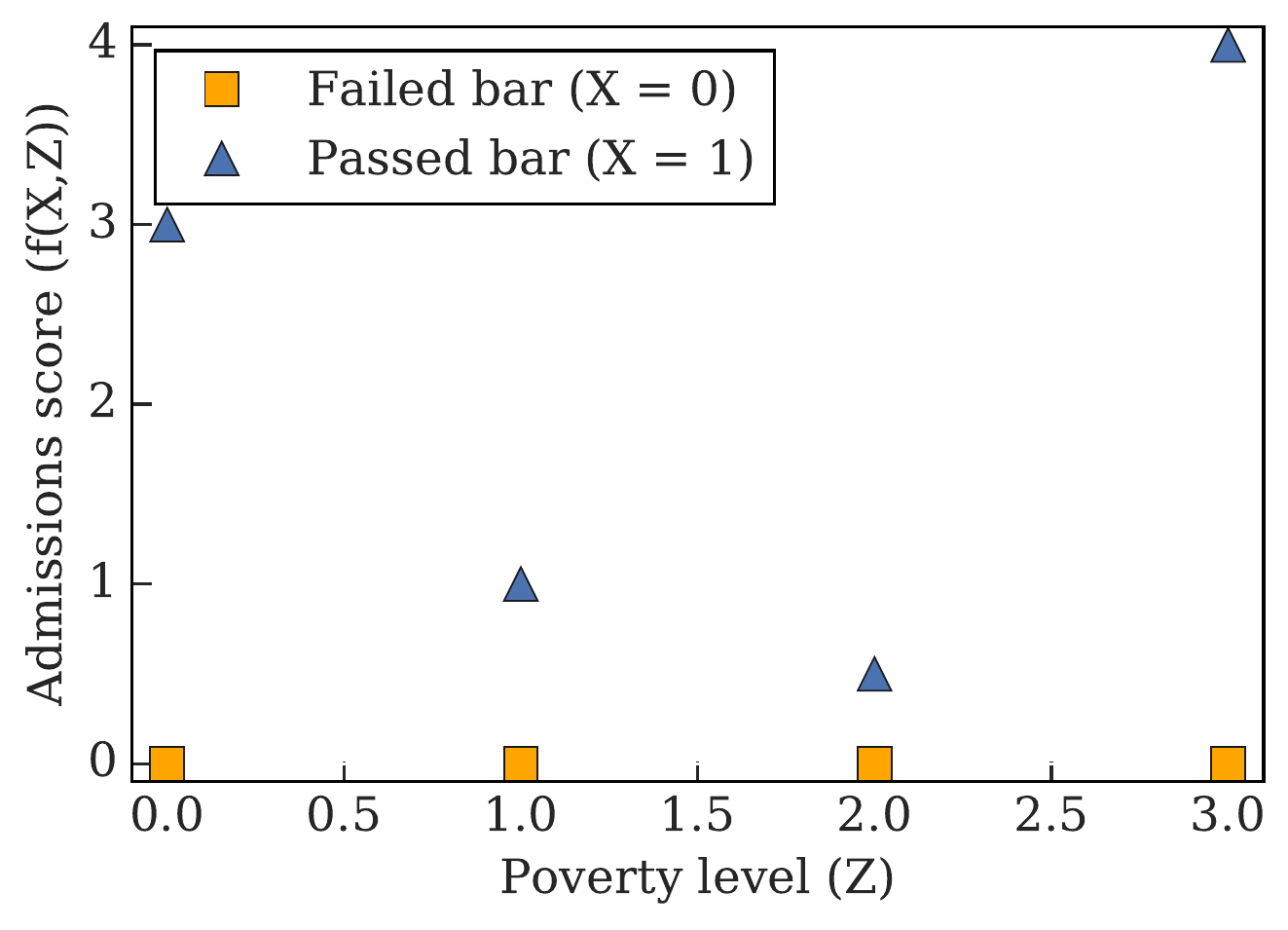} 
\caption{Nonmonotonic admissions scores for Counterexample \ref{app:moreunfaircounter}.}
\label{fig:nomoncounter} 
\end{figure}

While Lemma \ref{lem:moreunfair} shows that projecting a function onto monotonicity constraints cannot increase the \textit{average} \textit{one-sided statistical parity} violation, it can increase violations \textit{in the worst case}. Consider a continuation of the example from \ref{app:moncounter}, but this time let $f(X,Z)$ be defined by Fig. \ref{fig:nomoncounter}, and let $\tilde{f}(X,Z)$ be defined by Fig. \ref{fig:moncounter}. In this case, Fig. \ref{fig:moncounter} is the monotonic projection of Fig. \ref{fig:nomoncounter}. Then the worst case \textit{statistical parity} violation for the monotonic projection $\tilde{f}$ is \textit{higher} than the worst case \textit{statistical parity} violation for the non-monotonic $f$:
\begin{align*}
    &E[\tilde{f}(X,Z) | Z = 1] - E[\tilde{f}(X,Z) | Z = 2] \\
    &= 1.5(0.9) - 1.5(0.9) \\
    &= 1.2
\end{align*}
\begin{align*}
    &E[f(X,Z) | Z = 1] - E[f(X,Z) | Z = 2] \\
    &= 1.0(0.9) - 0.5(0.9) \\
    &= 0.85
\end{align*}

For a given pair $j,k$, as long as $\tilde{f}(x,j) \leq f(x,j)$ and $\tilde{f}(x,k) \geq f(x,k)$, then the violation $$R_f(j,k) =  E[f(X,Z) | Z = j] - E[f(X,Z) | Z = k]$$ will not be worse for the monotonic projection $\tilde{f}$: $R_{\tilde{f}}(j,k) \leq R_f(j,k)$. Lemma \ref{lem:moreunfair} holds because the inequalities $\tilde{f}(x,j) \leq f(x,j)$ and $\tilde{f}(x,k) \geq f(x,k)$ hold for $j = z_{(1)}$ and $k = z_{(|\mathcal{Z}|)}$, but this counterexample exists because those inequalities do not necessarily hold for any other pairs $j,k$ in between.

\section{Tradeoff between likelihood ratios in Lemma \ref{lem:binary}}\label{app:tradeoff}
The bound in Lemma \ref{lem:binary} contains two likelihood ratios: $\frac{p_{X|Z=j}(x)}{p_{X|Z=k}(x)}$ and $\frac{p_{X|\hat{Y}=1,Z=k}(x)}{p_{X|\hat{Y}=1,Z=j}(x)}$. When the first likelihood ratio is low, the second inverse likelihood ratio may be high. For example, suppose $Z$ is an individual's poverty level ($j$ being low poverty and $k$ being high poverty), $X$ is the number of extracurricular activities the individual is involved in, and $\hat{Y}=1$ means the individual is accepted into university. Suppose all individuals with above a certain number of extracurricular activities is accepted. Then the first likelihood ratio could be low when the number of extracurricular activities $X$ is low. Similarly, the likelihood that a high poverty individual accepted into university has a low number of extra curricular activities is probably also higher than the likelihood that a low poverty individual accepted into university has a low number of extracurricular activities. This implies that the second inverse likelihood ratio would be high, thus trading off with the first likelihood ratio.

\section{Further Analysis of Law School Admissions Experiments} \label{app:law}

\begin{figure}[h!]
\begin{tabular}{c}
Students that Passed the Bar Exam \\
\includegraphics[width=0.45\textwidth]{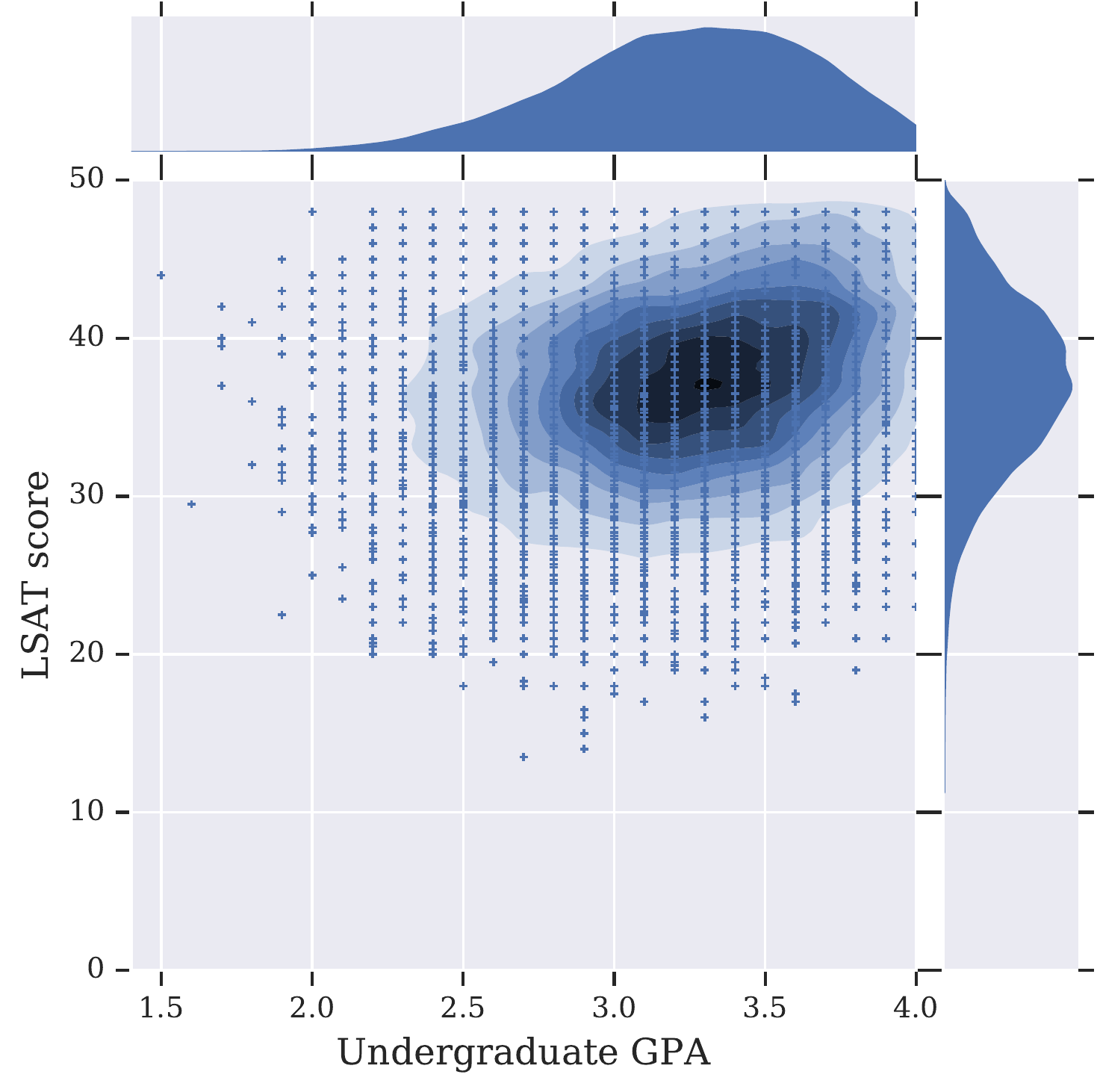} \\
Students that Failed the Bar Exam \\
\includegraphics[width=0.45\textwidth]{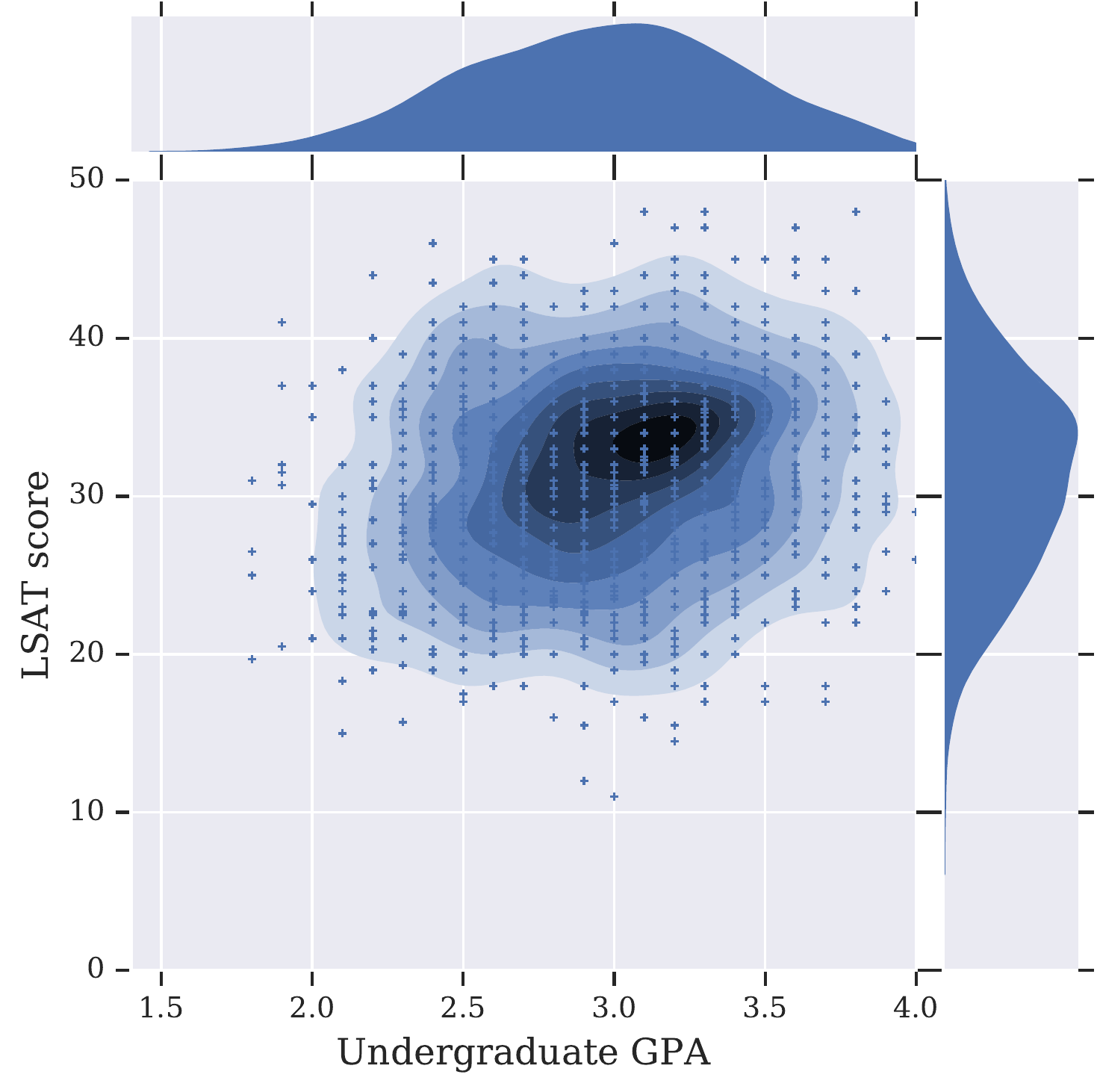}
\end{tabular}
\caption{Distribution over the full Law School Admissions dataset of undergraduate GPA and LSAT score students for students that passed the bar exam \textit{(top)} and students that failed the bar exam \textit{(bottom)}.} The dataset consists of $94.86\%$ students that passed the bar exam.
\label{fig:law_data} 
\end{figure}

Figure \ref{fig:law_data} shows the distribution of the LSAT scores, undergraduate GPA, and bar exam outcomes. Examples where the bar exam outcome was missing were omitted in our experiments.

\section{Further Analysis of Funding Proposals Experiments}\label{app:donors}

\begin{figure}[h!]
\centering
\includegraphics[width=0.49\textwidth]{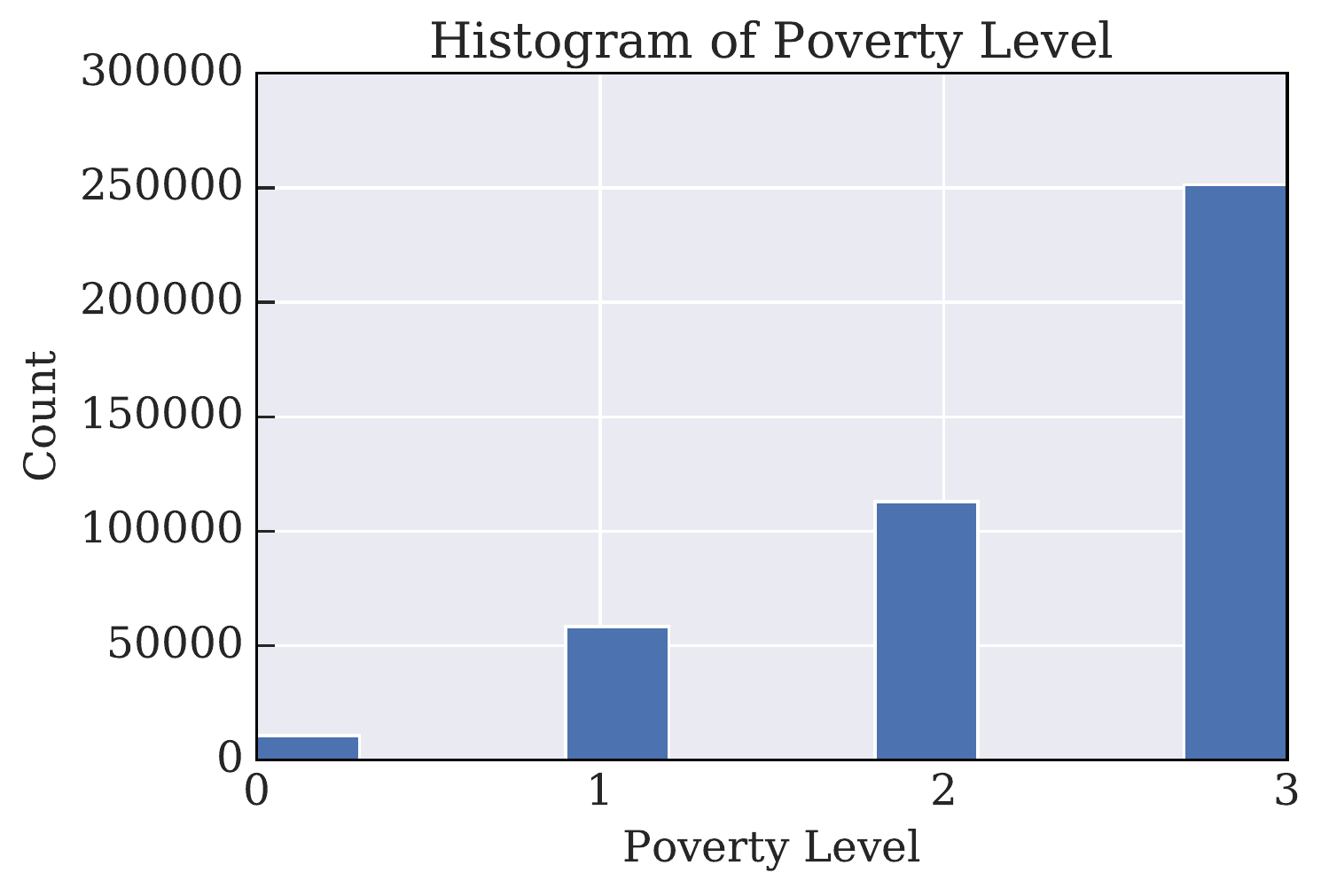}
\caption{Histogram of the poverty level feature from the Funding Proposals dataset. 0 represents lowest poverty and 3 represents highest poverty.}
\label{fig:histogram} 
\end{figure}

Figure \ref{fig:histogram} gives a histogram of the four different poverty levels, which are ordinal with level 3 being the most impoverished.

\begin{figure}[h!]
\centering
\begin{tabular}{c}
Funding Proposals: Training Examples\\
\includegraphics[width=0.5\textwidth]{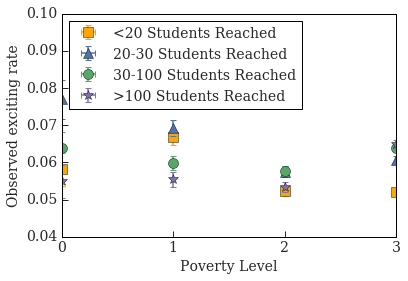}\\
\\
Funding Proposals: Unconstrained Model Predictions \\
\includegraphics[width=0.5\textwidth]{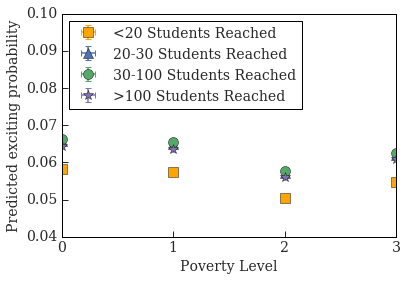}\\
\\
Funding Proposals: Monotonic Model Predictions \\
\includegraphics[width=0.5\textwidth]{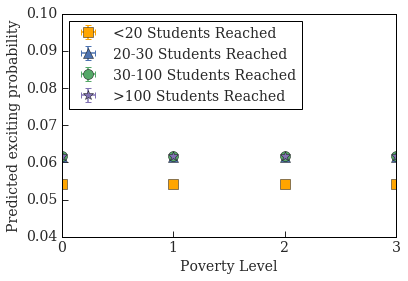}\\
\end{tabular}
\caption{\textit{(top)} Plot of the observed rate of exciting projects (mean number of exciting projects) as a function of each project's poverty level and number of students reached. Error bars show the standard deviation.  \textit{(middle)} Unconstrained model predictions. \textit{(bottom)} Shape-constrained model predictions. }
\label{fig:donors3} 
\end{figure}

Figure \ref{fig:donors3} \emph{(top)} shows the training examples' average number of exciting projects, where the error bars show the standard error of the mean.  The poverty level feature ranges from 0 to 3, with 0 denoting low poverty and 3 denoting the highest poverty level. For ease of visualization, we show the quartiles of the students-reached feature.

Figure \ref{fig:donors3} \emph{(middle)} shows the predicted probability that a project is exciting for a GAM  model without the proposed ethical constraints. The model  gives lower scores to  poverty level 2 (poorer schools) than to poverty level 1 (richer schools) for every quartile of students reached.  The model also gives higher scores for project that reach 30-100 students tahn to projects that reach 100+ students.

Figure \ref{fig:donors3} \emph{(bottom)} shows that training with an ethical monotonicity shape constraint works: at the same poverty level, projects that affect more students are given a higher score. For the same quartile of students reached, the score also does not decrease for higher poverty levels.

\end{document}